\documentclass[10pt,journal,compsoc]{IEEEtran}

\ifCLASSINFOpdf

\else

\fi

\usepackage{xpatch}
\usepackage[dvipsnames]{xcolor}

\usepackage{graphicx}
\usepackage{comment}
\usepackage{booktabs}
\usepackage{multirow}

\usepackage{amsmath} 
\usepackage{amssymb} 
\usepackage{color}
\usepackage{subfigure}

\usepackage{ntheorem}
\usepackage{mathtools}
\usepackage{enumitem}
\usepackage{url}
\usepackage{soul}
\usepackage[cmintegrals]{newtxmath}

\usepackage[capitalise]{cleveref}
\usepackage{tabu}

\definecolor{mygray}{gray}{0.95}
\definecolor{black}{gray}{0.5}

\theoremseparator{:}
\newcommand{\etal}{{et al}.}
\newcommand{\ie}{i.e.}
\newcommand{\eg}{e.g.}

\begin{document}

\title{CoDiNet: Path Distribution Modeling with Consistency and Diversity for Dynamic Routing}

\author{Huanyu~Wang,
		Zequn~Qin,
		Songyuan~Li,
        and~Xi~$\textup{Li}^{\dagger}$
\IEEEcompsocitemizethanks{\IEEEcompsocthanksitem H. Wang, Z. Qin, S. Li are with the College of Computer Science and Technology, Zhejiang University, Hangzhou, Zhejiang, China, 310007. (e-mail: huanyuhello@zju.edu.cn, qinzequn@zju.edu.cn, leizungjyun@zju.edu.cn).
\IEEEcompsocthanksitem X. $Li^{\dagger}$ is with the College of Computer Science and Technology, Zhejiang University, Hangzhou, China, 310007. (email: xilizju@zju.edu.cn). 
\IEEEcompsocthanksitem The corresponding author of this paper is Prof. Xi Li.
}
}

\markboth{IEEE Transactions on Pattern Analysis and Machine Intelligence}%
{Shell \MakeLowercase{\textit{et al.}}: Bare Advanced Demo of IEEEtran.cls for IEEE Computer Society Journals}

\IEEEtitleabstractindextext{

\begin{abstract}
    Dynamic routing networks, aimed at finding the best routing paths in the networks, have achieved significant improvements to neural networks in terms of accuracy and efficiency. In this paper, we see dynamic routing networks in a fresh light, formulating a routing method as a mapping from a sample space to a routing space. From the perspective of space mapping, prevalent methods of dynamic routing did not take into account how routing paths would be distributed in the routing space.
    Thus, we propose a novel method, termed CoDiNet, to model the relationship between a sample space and a routing space by regularizing the distribution of routing paths with the properties of consistency and diversity. 
    In principle, the routing paths for the self-supervised similar samples should be closely distributed in the routing space. 
    Moreover, we design a customizable dynamic routing module, which can strike a balance between accuracy and efficiency. When deployed upon ResNet models, our method achieves higher performance and effectively reduces average computational cost on four widely used datasets. 
\end{abstract}

}

\maketitle

\IEEEdisplaynontitleabstractindextext

\IEEEpeerreviewmaketitle

\ifCLASSOPTIONcompsoc
\IEEEraisesectionheading{
\section{Introduction}\label{sec:introduction}}
\else
\section{Introduction}
\label{sec:introduction}
\fi

\begin{figure}[t]
	\centering
	\includegraphics[width=1.0\linewidth]{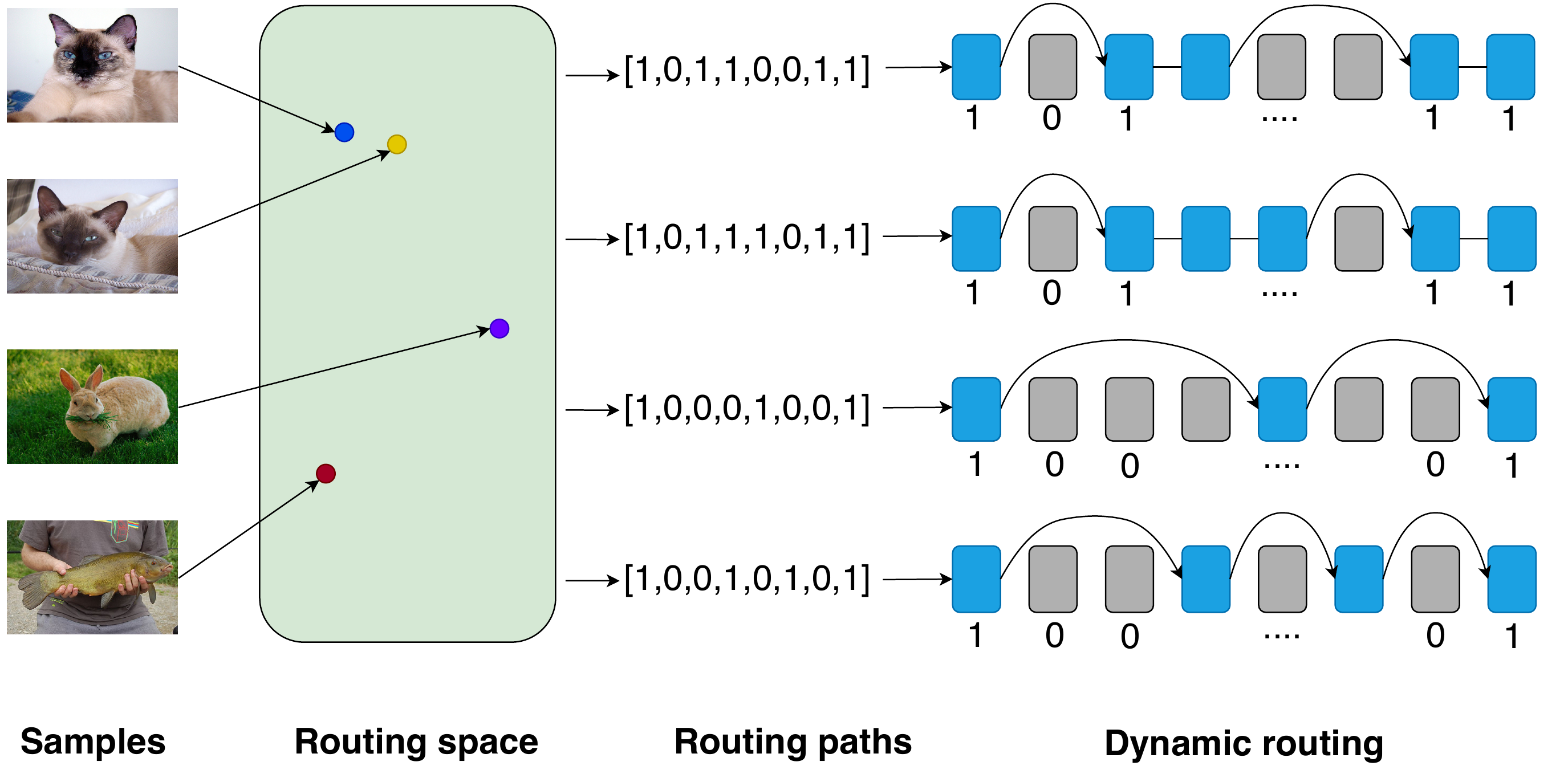}
	\caption{Illustration of dynamic routing. A dynamic routing network is a mapping from a sample space to a routing space. Each routing path consists of a sequence of to-be-run and to-be-skipped blocks. The dynamic routing model walks through the to-be-run blocks. Best viewed in color.
	}
	\label{fig.mapping}
\end{figure}

\IEEEPARstart{D}{ynamic} routing is a sample-adaptive inference mechanism for neural networks. At inference time, only part of a dynamic routing network would be activated for each sample, which is aimed at reducing computational cost with little performance compromised~\cite{veit2018convolutional, wang2018skipnet, wu2018blockdrop, almahairi2016dynamic}. 
In essence, dynamic routing can be considered as a mapping from a sample space to a routing space. As shown in \cref{fig.mapping}, when samples are presented to a dynamic routing model, they are mapped into a routing space. Each sample's routing path in the routing space can be represented as a binary vector, consisting of a sequence of to-be-run and to-be-skipped blocks. The model walks through the to-be-run blocks. 

\begin{figure*}[t]
	\centering
	\includegraphics[width=\linewidth]{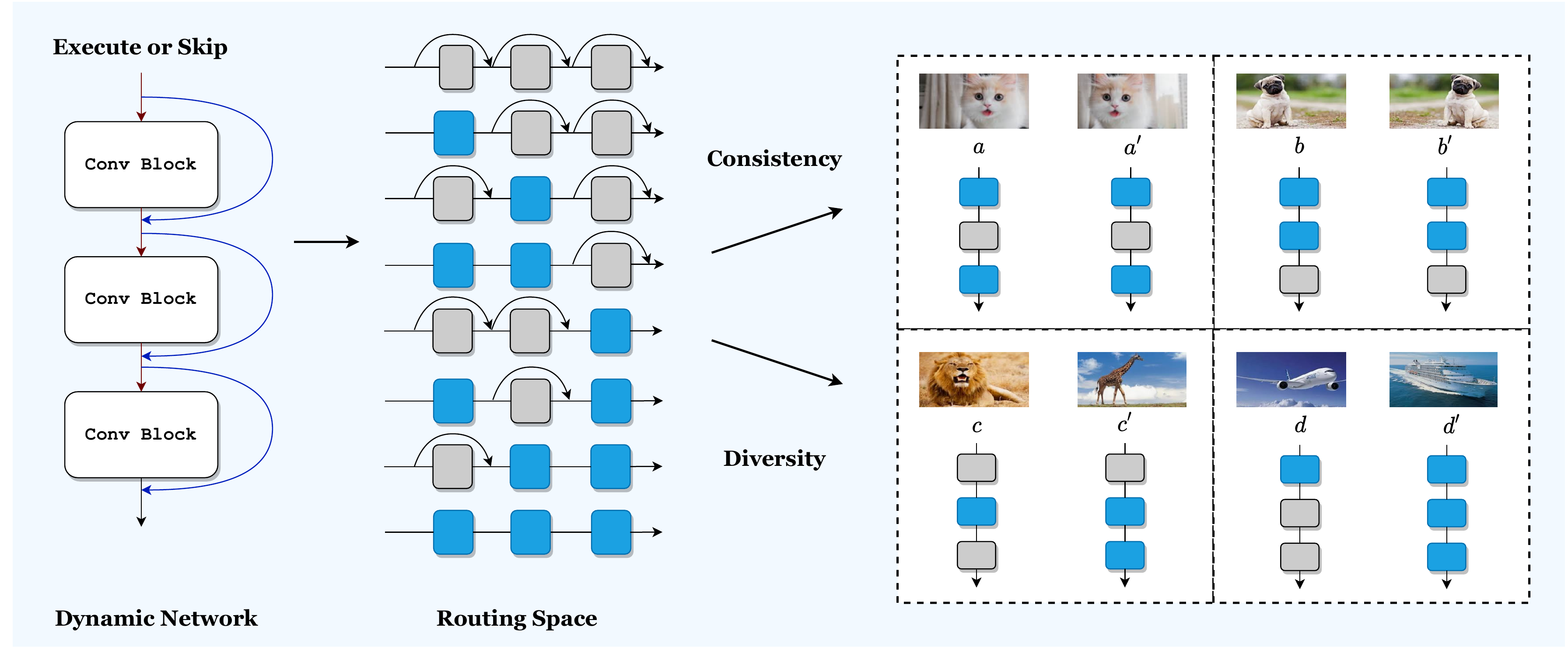}
	\caption{Illustration of our motivation. \textbf{Left:} Schematic diagram of a dynamic routing network, in which each layer can be either executed or skipped. \textbf{Middle:} A demonstration of the routing space. All potential routing paths compose a binary routing space. \textbf{Right:} Routing paths for similar images should be the same or similar, \eg, $a$ and $a'$, $b$ and $b'$, while routing paths for dissimilar images should be different, \eg, $c$ and $c'$, $d$ and $d'$. Best viewed in color.}
	\label{fig_idea}
\end{figure*}

From the perspective of space mapping, how inference
paths should be distributed in the routing space remains relatively unexplored.
Since dynamic routing is related to two spaces, i.e., a sample space and a routing space, we focus on this question: \textit{how do we model the relationship between the two spaces?}
We expect that the distance between similar samples should be close in the routing space. Otherwise, the distance should be far. 
Moreover, \textit{what kind of samples are similar?} We utilize the self-supervised similarity and encourage the self-supervised augmentations walking through close routing paths. Since the images obtained by self-supervised augmentations remain semantically and visually unchanged, their routing paths should be close in the routing space. In comparison, images belong to the same semantic class might be significantly different due to different background, object layout, and color. The feature extraction patterns of these samples are different so that it is infeasible to utilize close routing paths for such visually different images. 
As shown in \cref{fig_idea}, images $a$ and $a'$, images $b$ and $b'$ are similar to each other, while $c$ and $c'$, $d$ and $d'$ are dissimilar. Therefore, the routing paths of images $a$ and $a'$, images $b$ and $b'$ should be the same or similar, while routing paths of images $c$ and $c'$, $d$ and $d'$ should be different. 

To this end, we explicitly model the relationship between the sample space and the routing space, therein establishing the connection between samples and routing paths. We propose a novel dynamic routing method, termed CoDiNet, to regularize the path distribution in a routing space with the properties of consistency and diversity, based on the aforementioned space mapping. 
Firstly, the consistency regularization makes augmentations of the same sample have similar feature activations, thus forming a specific routing paths for specific samples. Parameters on the specific routing path are consistently stimulated by similar samples, which is favorable to parameter sharing among similar samples and robustness of the network. 
At the same time, the diversity regularization makes the routing paths generated by dynamic routing more diverse, which strengthens the exploration of the network. It is evident that the more routing paths are used, the more capacity of the network is utilized. 


Computational cost is another important issue because the computational ability of different platforms varies considerably. For instance, the inference speed of ResNet-50~\cite{he2016deep} on GTX 1080ti \textit{(30fps)} is much faster than that on Maxwell TitanX \textit{(18fps)} at a resolution of $224\times224$. The average cost of dynamic routing networks should be customizable, when the networks are applied to platforms with different computational budgets.
To this end, we propose a differentiable computational cost loss to optimize the average computational cost of the model and make the computational budgets customizable.

The main contributions of CoDiNet can be summarized into three parts:
\begin{itemize} [leftmargin=*]
	\item We explicitly model the relationship between the sample space and the routing space, therein establishing the connection between samples and routing paths.
	\item We propose a novel consistency-and-diversity regularized optimizing method modeling the relationships between the two spaces, which makes routing paths optimizable.
	\item We design a customizable dynamic routing module and achieve state-of-the-art results in terms of computational cost reduction and performance.
\end{itemize}

\begin{figure*}[h]
	\subfigure[Dynamic routing]{
		\label{sfig:dynamic_routing}
		\centering
		\includegraphics[width=.45\linewidth, height=1.5in]{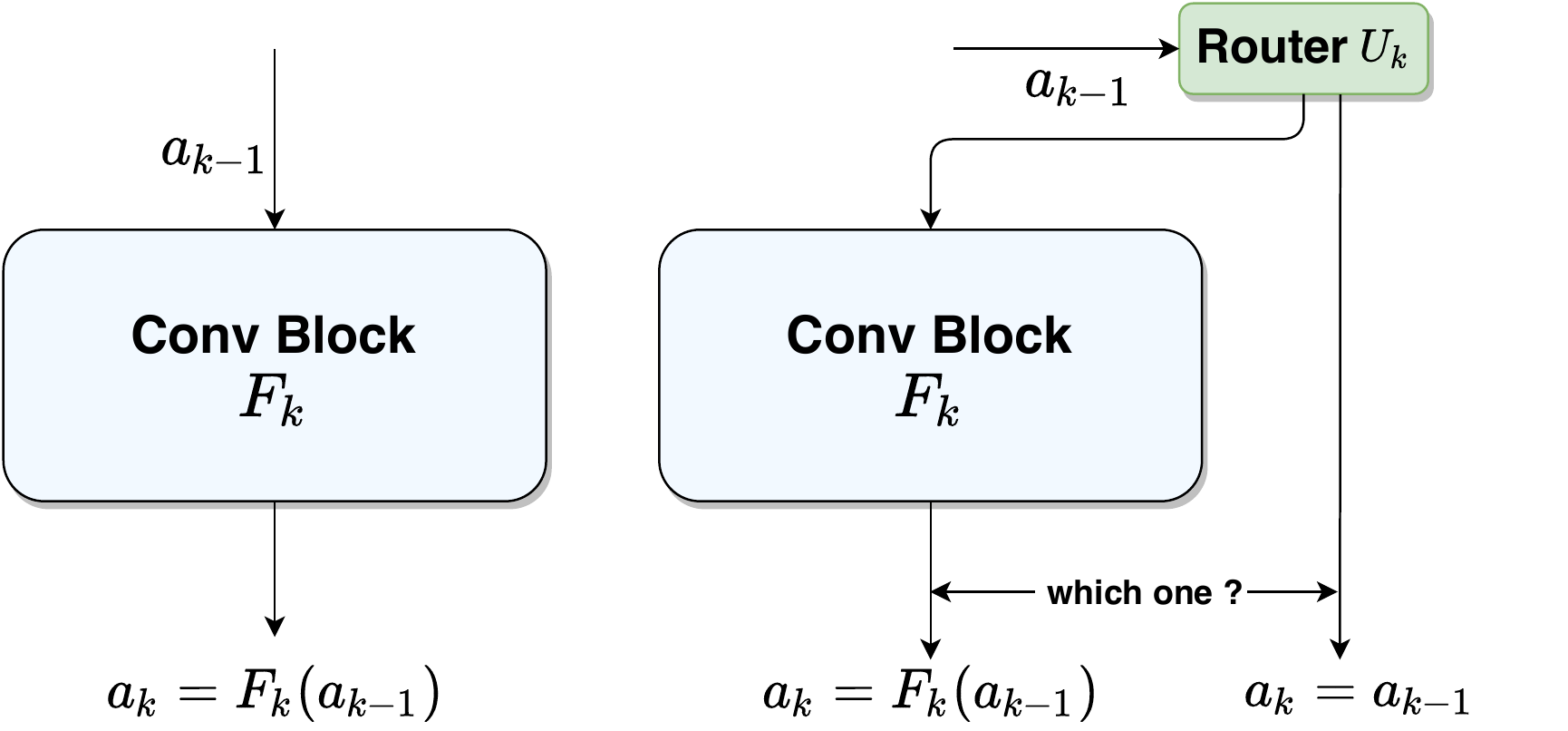}
	}
	\subfigure[Our router]{
		\label{sfig:routing_module}
		\centering
		\includegraphics[width=.53\linewidth, height=1.5in]{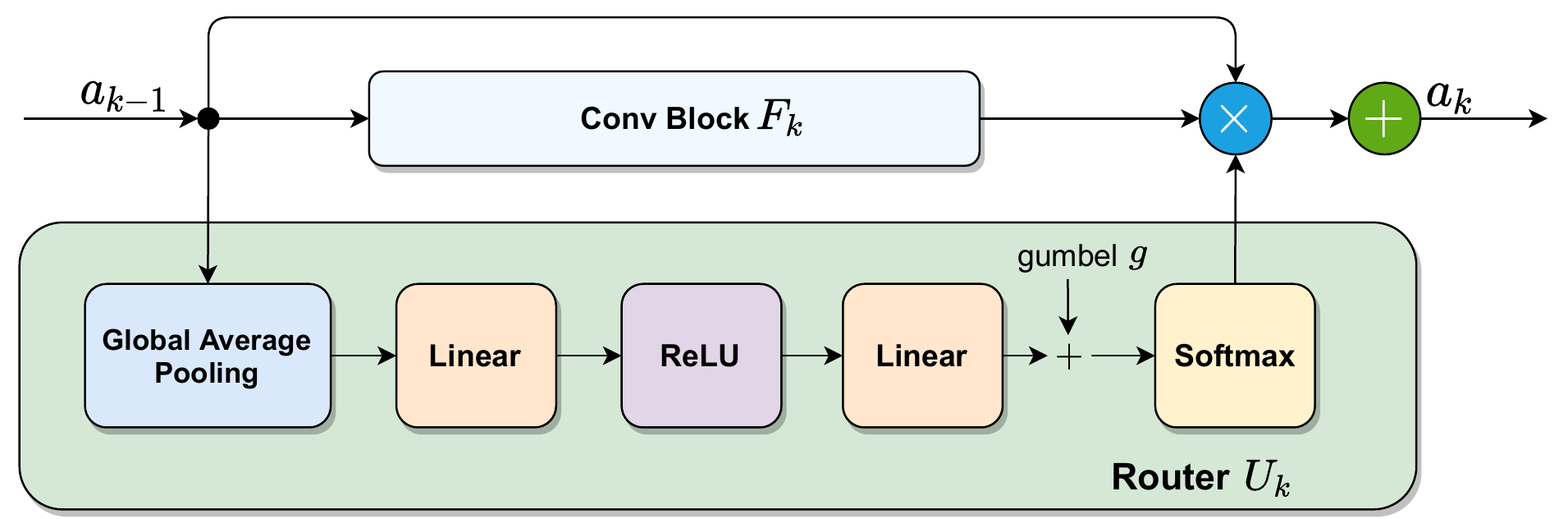}
	}
	\caption{Illustration of dynamic routing and our router. (a) Comparison between static inference and dynamic routing. In the dynamic routing network, whether the current convolution block would be executed depends on the output of the previous block. 
	(b) Illustration of the structure of our router. The cost of our router is negligible compared with a convolution block.}
	\label{fig.route_structure}
\end{figure*}

\section{Related Works}\label{sec:related}

In this section, we revisit relevant methods and divide them into three categories: dynamic routing networks, early prediction networks, and model compression methods. Dynamic routing and early prediction are two typical approaches to dynamic inference. 
The former focus on skipping unnecessary units at inference time, while the latter is characterized by multiple exits. Model compression methods are also popular for cost reduction with static inference. 

\subsection{Dynamic Routing Networks}
Layer dropping has long been used as a regularization technique in neural networks, e.g., DropConnection~\cite{wan2013regularization} and Dropout~\cite{JMLR:v15:srivastava14a}. Veit \etal~\cite{EnsemblesShallow} found that only short paths of deep residual networks are needed. Motivated by this, dynamic routing networks have emerged as a promising technique to skip blocks or layers at inference time for acceleration~\cite{veit2018convolutional, wang2018skipnet, wu2018blockdrop, almahairi2016dynamic, su2020dynamic}. Specifically, ConvNet-AIG~\cite{veit2018convolutional} proposed a convolutional network that adaptively defines its inference graph conditioned on the input images. it proposes a router to make the execution decision for each convolutional block. SkipNet~\cite{wang2018skipnet} introduced a method with LSTM gate-ways to determine whether the current block would be skipped or not. Besides, BlockDrop~\cite{wu2018blockdrop} adopted an extra policy network to sample routing paths from the whole routing space to speed up ResNets' inference. Slimmable Nets~\cite{yu2018slimmable} intended to train a model to support multiple widths to fit different computational constraints. Recursive network~\cite{guo2019dynamic} proposes to execute a convolutional layer multiple times. RNR~\cite{rao2018runtime} models the dynamic process as a Markov decision process and uses reinforcement learning for training.
Spatial dynamic convolutions for fast inference were proposed in~\cite{verelst2020dynamic, yu2019universally, sun2020computation, xie2020spatially}. Multi-scale networks were introduced in~\cite{huang2018multi, yang2020resolution}. They learn easy samples at low resolutions, while hard samples at high resolutions. Channel-based dynamic routing methods~\cite{su2020dynamic, jordao2020discriminative} were introduced as well. 
Recently, various dynamic methods with different kinds of selection have been proposed. 
Multi-kernel methods~\cite{Chen_2020_CVPR, chen2020dynamic} select different CNN kernels for better performance. Recursive network~\cite{guo2019dynamic} are introduced to reuse the networks. 

At training time, dynamic routing  models are prone to early convergence to suboptimal states. To deal with the issue, SkipNet~\mbox{\cite{wang2018skipnet}} uses multiple training stages, Blockdrop~\mbox{\cite{wu2018blockdrop}} uses curriculum learning, dynamic conv~\mbox{\cite{verelst2020dynamic}} uses annealing, sparsity network~\mbox{\cite{sun2020computation}} uses a non-conditional pre-training.
In comparison, we focus on how to learn proper paths. To this end, we model the relation between samples and their routing paths explicitly and optimize the routing paths directly, achieving a more stable dynamic routing model.

\subsection{Early Prediction Networks}
While a dynamic routing network has only one exit, an early prediction network is characterized by multiple exits.
In an early prediction network, the network exits once the criterion for a certain sample is satisfied.
Traditional methods~\cite{reyzin2011boosting, hu2014efficient} applied heuristic and greedy algorithms to reduce the executed layers. 
BranchyNet~\cite{teerapittayanon2016branchynet} proposed a multiple-branch framework by attaching fully connected layers to intermediate layers of the backbone. 
ACT~\cite{graves2016adaptive} proposed a halting unit for a recurrent neural network (RNN) to realize early prediction. Following ACT, SACT~\cite{figurnov2017spatially} proposed a CNN-based early prediction network, adopting a stopping unit for each point on feature maps.
Since then, early prediction frameworks have been widely used in classification for efficient inference. 

Considering multi-scale inputs, MSDN~\cite{Huang2017MultiScaleDN} introduced early-exit branches based on DenseNet~\cite{huang2017densely}. According to the allowed time budget, McIntosh et al.~\cite{mcintosh2018recurrent} proposed an RNN architecture to dynamically determine the exit. Li \etal~\cite{li2019improved} proposed a self-distillation mechanism to supervise inter-layer outputs with deeper layers. Instead of bypassing residual units, DCP~\cite{gao2018dynamic} generated decisions to save the computational cost for channels. Hydranets~\cite{Hydranets} proposed to replace the last residual block with a Mixture-of-Experts layer. Recently, methods have been adopted to other applications, such as action recognition~\cite{hussein2020timegate, meng2020ar} and object detection~\cite{zhang2019slimyolov3}. 
Our method belongs to dynamic routing networks. Thus, our method does not have multiple exits as early prediction networks do. We also compare our method with early prediction networks in \cref{sssec:sotas}.

\subsection{Model Compression Methods}
Compression methods are proposed for high-performance models on platforms with limited computational resources. Knowledge distillation~\cite{hinton2015distilling, chen2017learning, chen2018distilling, yu2018nisp}, low-rank factorization~\cite{ioannou2015training, tai2015convolutional, jaderberg2014speeding}, and quantization~\cite{han2015deep, wu2016quantized, polino2018model} have been widely used to compress the structures and to prune the parameters of neural networks. Besides, recent researches tend to prune unimportant filters or features~\cite{li2016pruning, he2017channel, luo2017thinet, wen2016learning, huang2018condensenet} to compress or speed-up the model. They identify ineffective channels or layers by examining the magnitude of the weight or activation. The relatively ineffective channels and layers are pruned from the model. Then the pruned model is finetuned to mitigate the accuracy loss. With the iteration of pruning unnecessary parts and then finetuning the model, computational cost and model size can reduce effectively. Perforated CNN~\cite{Perforated} speeds up the inference by skipping the computations at fixed spatial locations. In addition, Neural Architecture Search provides other technique plans achieving low-cost models including MnasNet \cite{tan2019mnasnet}, ProxylessNAS \cite{cai2018proxylessnas}, EfficientNet \cite{tan2019efficientnet}, and FbNet \cite{wu2019fbnet}. In contrast to this line of work where the same amount of computation is applied to all samples, we focus on efficient inference by dynamically choosing a series of blocks to be executed conditioned on the input.

\section{Methods}

\begin{figure*}[t]
	\centering
	\includegraphics[width=1.0\linewidth]{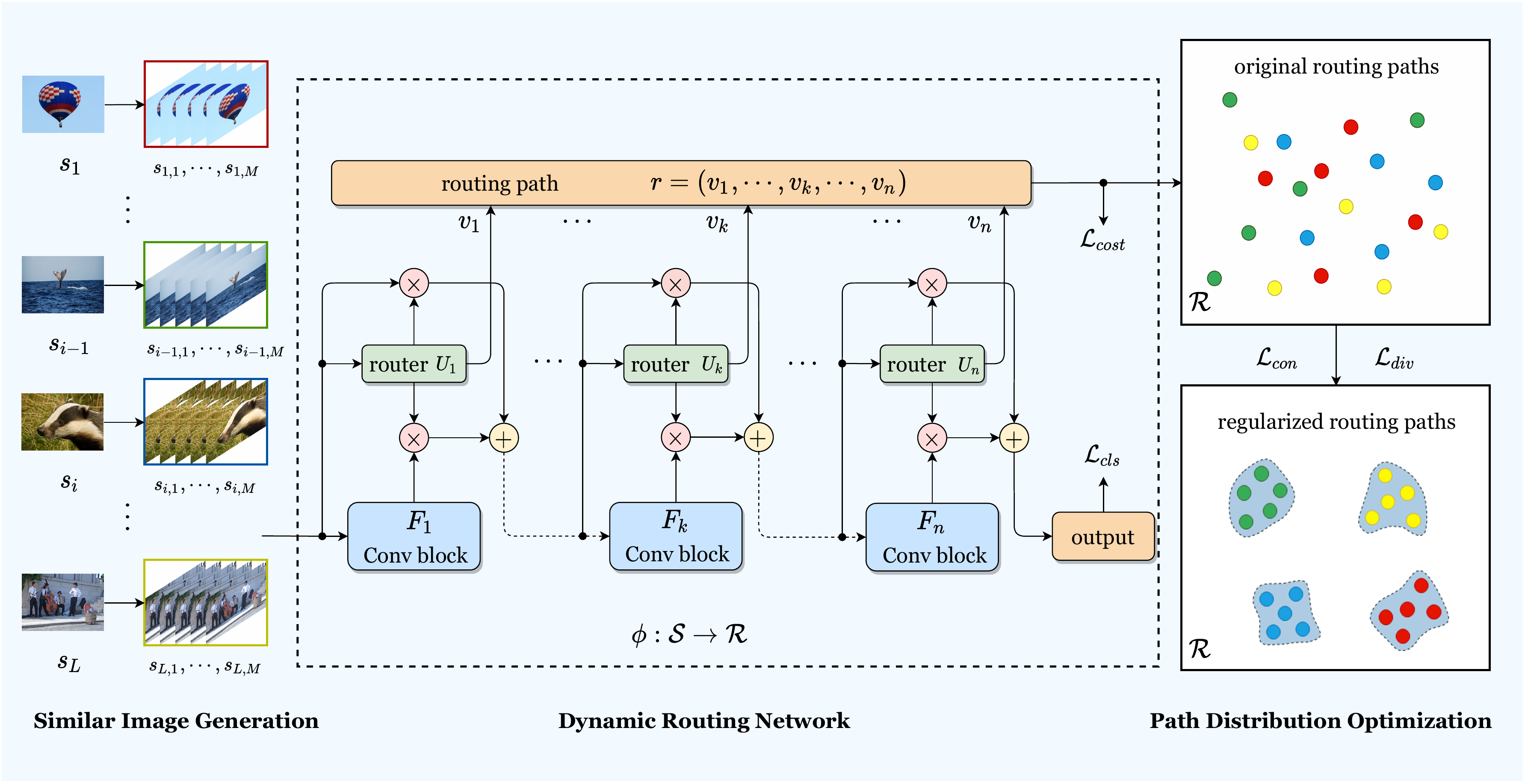}
	\caption{An overview of CoDiNet. \textbf{Similar Image Generation}: Each sample is augmented randomly and get several similar augmentations. \textbf{Dynamic Routing Network}: Through a dynamic routing network, we get routing paths and prediction for each sample. \textbf{Path Distribution Optimization}: We optimize the path distribution with the consistency loss $\mathcal{L}_{con}$ and the diversity loss $\mathcal{L}_{div}$. Note that the augmentation methods are commonly used in classification, which are cropping and horizontally flipping. Best viewed in color.
	}
	\label{fig.min_max}
\end{figure*}

In this section, we illustrate our method CoDiNet in detail. 
First, we formulate dynamic routing as a space mapping and introduce the basics of dynamic routing accordingly.
Second, we show how to model the relationship between the two spaces with the regularization of consistency and diversity. Third, we design a training strategy to make our dynamic routing network adaptive to different computational budgets. Finally, we illustrate our process of training and inference. 
For convenience, \cref{tab:notation} summarizes the notations.

\begin{table}[tb]
	\centering
	\caption{Notations}
	\label{tab:notation}
	\vspace{-1em}
	\begin{tabu} to 0.49\textwidth {X[c]X[4]}
	\toprule
	$\mathcal{S}$ 	& a sample space\\
	$s_i$ 			& a sample \\
	$\mathcal{R}$ 	& a routing space\\
	$r_i$ 			& the routing path for $s_i$\\
	$\mathcal{\phi}$& an $n$-block network\\
	$F_k$		 	& the $k$-th block of $\mathcal{\phi}$\\
	$a_k$		 	& the input of the $(k+1)$-th block of $\mathcal{\phi}$\\
	$U_k$		 	& the router for the $k$-th block \\
	$u_k$		 	& the execution decision of the $k$-th block \\
	$v_k$		 	& the relaxation of $u_k$ \\
	\midrule
	$m_c$		 	& the margin for consistency \\
	$m_d$		 	& the margin for diversity \\
	$\mathcal{L}_{con}$		 	& the loss function for  consistency \\
	$\mathcal{L}_{div}$		 	& the loss function for  diversity	\\
	$\mathcal{L}_{cost}$		& the loss function for  customizable dynamic routing\\
	$\alpha$		& the hyper-parameter for $\mathcal{L}_{con}$ \\
	$\beta$			& the hyper-parameter for $\mathcal{L}_{div}$  \\
	$\gamma$		& the hyper-parameter for $\mathcal{L}_{cost}$  \\
	\bottomrule
	\end{tabu}
\end{table}

\subsection{Dynamic Routing as a Space Mapping}
\subsubsection{Routing Space}
We see dynamic routing as a mapping from a sample space to a routing space. A sample space $\mathcal{S}$ is a set of samples, and a routing space $\mathcal{R}$ is the set of all the possible routing paths of a dynamic routing network $\mathcal{\phi}$. In this way, dynamic routing can be considered as a mapping $\mathcal{\phi}:\mathcal{S} \rightarrow \mathcal{R}$. That is, for each sample $s_i\in\mathcal{S}$, its routing path $r_i \in \mathcal{R}$ in a dynamic routing network $\mathcal{\phi}$ is
\begin{equation}
	r_i = \mathcal{\phi}(s_i).
\end{equation}

A routing path consists of a sequence of to-be-skipped and to-be-run blocks. In this paper, we use the same block as in ResNet~\cite{he2016deep}. Let $\mathcal{\phi}$ be an $n$-block network and $u_k \in \{0,1\}$  be the execution decision for the $k$-th block, where $0$ stands for to-be-skipped and $1$ stands for to-be-run.  Then, a routing path for the sample $s_i$ is a concatenation of decisions for all the blocks, \ie, $r_i = (u_1, u_2, \cdots, u_n)$.
Hence, the routing space of $\mathcal{\phi}$  is $\mathcal{R}={\{0, 1\}^{n}}$, which contains $2^n$ routing paths in total.

\subsubsection{Routers}
For each block in the network $\mathcal{\phi}$, there is a router used to decide whether the block should be executed for a specific sample.  Let  $F_{k}$ be the $k$-th block of $\mathcal{\phi}$, $a_{k-1}$ be the input of $F_{k}$, and $U_k$ be the router for the $k$-th block. 
Then, the dynamic routing result $a_k$ of the $k$-th block is defined as
\begin{equation}
	a_{k} = u_{k} \cdot F_{k}(a_{k-1}) + (1-u_{k}) \cdot a_{k-1},
\end{equation}
where the value of $u_k$ stands for the decision of $U_k$.

Routers are supposed to find the correct path while incurring a low computational cost.
To minimize the cost, we use a lightweight structure for each router, which only contains two fully connected layers, as shown in \cref{sfig:routing_module}.
First, the router $U_k$ gathers information from the block input $a_{k-1} \in \mathbb{R}^{W_k \times H_k \times C_k}$ across channels by global average pooling, which is written as
\begin{equation}
	z_k = \mathrm{gap}(a_{k-1}).
\end{equation}
Then, the fused features $z_k$ are processed by two fully connected layers sequentially. Let $\mathrm{W}_1 \in \mathbb{R}^{d \times C_k}$ and $\mathrm{W}_2 \in \mathbb{R}^{2 \times d}$ be the weights of $U_k$'s first layer and second layer respectively, where $d$ denotes the output dimension of the first layer. The router $U_k$ is defined as 
\begin{equation}
	U_k = \mathrm{W}_2 \circ \sigma(\mathrm{W}_1 \circ z_k),
\end{equation}
where $\sigma(\cdot)$ is an activation function, and $\circ$ denotes matrix multiplication. It is worth noting that $U_k$ is a two-element vector as a result. $U_k[0]$ is the first element of $U_k$, and $U_k[1]$ is the second element. In this way, the execution decision $u_k$  of the $k$-th block is calculated by
\begin{equation}
	u_k = \mathop{\arg\max}_j(U_k[j]), 
\end{equation}
where $u_k=0$ means to-be-skipped, and $u_k=1$ means to-be-run as a result.
The simple yet effective structure incurs less than one percent of the computational cost of a convolution block. 

\begin{figure*}[t]
	\centering
	\subfigure[Routing Space]{
		\centering
		\includegraphics[height=1.6in, width=0.23\linewidth]{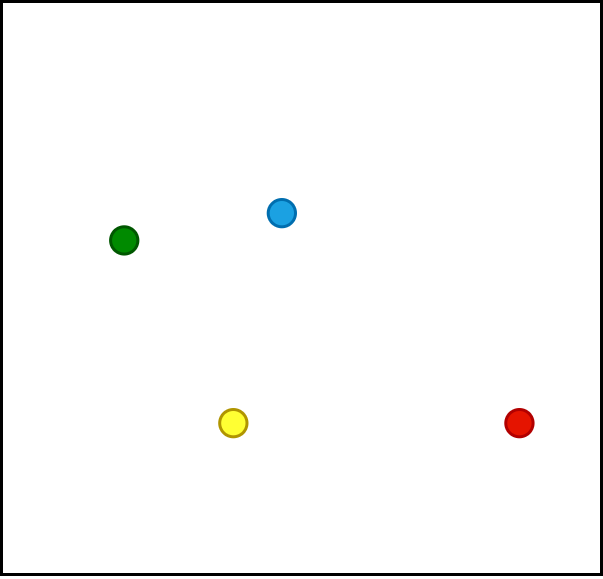}
		\label{sfg.insight_a}
	}
	\subfigure[Augmentation]{
		\centering
		\includegraphics[height=1.6in, width=0.23\linewidth]{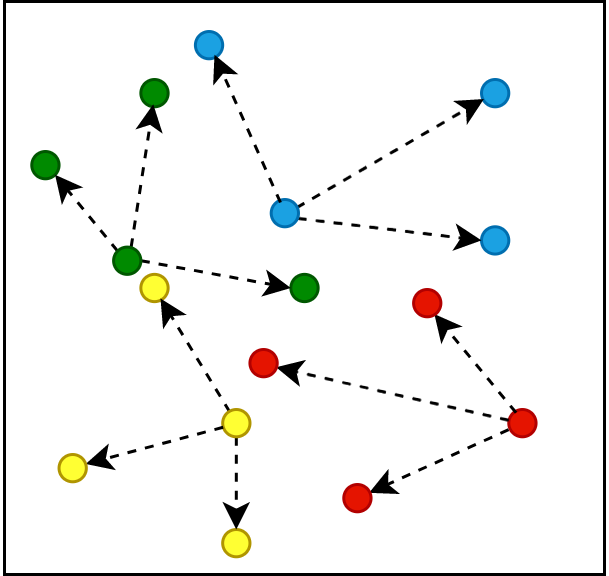}
		\label{sfg.insight_b}
	}
	\subfigure[Consistency]{
		\centering
		\includegraphics[height=1.6in, width=0.23\linewidth]{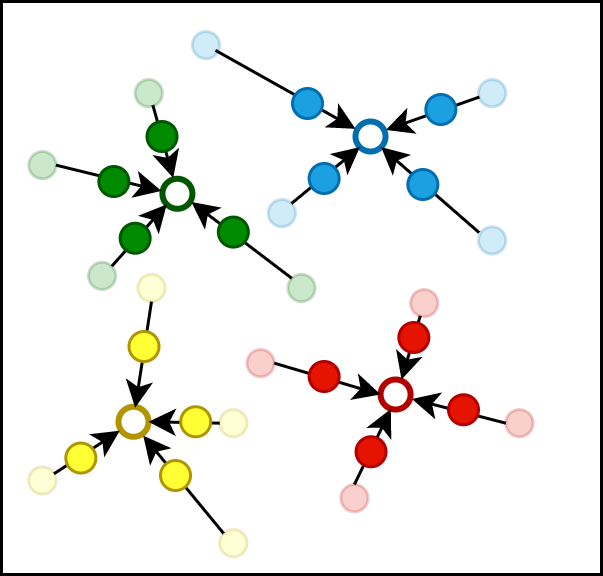}
		\label{sfg.insight_c}
	}
	\subfigure[Diversity]{
		\centering
		\includegraphics[height=1.6in, width=0.23\linewidth]{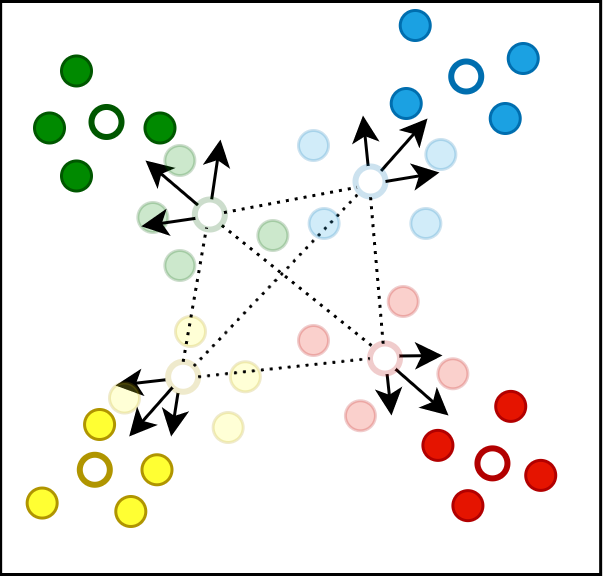}
		\label{sfg.insight_d}
	}
	\caption{Illustration of the optimization of path distribution. A solid point means a routing path of an image. A hollow circle means a routing path center of similar images. (a) Suppose we have four different input images whose paths distribute in the routing space. (b) With augmentations, we can get four similar instances of each image. (c) The consistency regularization makes routing paths of similar instances cluster around their center. (d) The diversity regularization disperses the center of dissimilar instances. Best viewed in color.
	}
	\label{fig.insight}
\end{figure*}
\subsubsection{Route Relaxation}
\label{sec.relaxation}
To make the binary routing decisions $u_1, u_2, \cdots, u_n$ optimizable in an end-to-end fashion, we utilize a continuous, differentiable relaxation function, Gumbel-Softmax \cite{gumbel1948statistical}, at the end of each router as shown in \cref{sfig:routing_module}. Gumbel-Softmax turns discrete values into continuous ones, enabling backpropagation. 
Let $g$ be the noise samples from a Gumbel distribution, and let $\mathcal{T}$ be the temperature which is fixed to $1$ in our experiments. The relaxation for $u_k$, called $v_k$, is calculated by
\begin{equation}
	v_k = \mathrm{softmax}(\frac{\log(U_k)+g}{\mathcal{T}})[1],
\end{equation}
where $\mathrm{softmax}(\cdot)[1]$ is the second element.
In this way, the dynamic routing result of the $i$-th block is relaxed by
\begin{equation}
	a_{k} = v_k \cdot F_{k}(a_{k-1}) + (1-v_k) \cdot a_{k-1}.
	\label{eq1}
\end{equation}
The routing path $r_i$ for an $n$-block network is also relaxed by $r_i = (v_1, v_2, \cdots, v_n)$, where each element is in $[0,1]$.
It is worth noting that a block in $\mathcal{\phi}$ is either run or skipped at inference time. We will  describe it in \cref{sssec:inference}.

Next, we propose to regularize routing paths in the routing space.

\subsection{Consistency and Diversity Regularization}
\label{loss}
We expect that the routing paths of similar samples should be consistent in the routing space. Otherwise, the routing paths should be diverse. In real scenarios, the similarity of different images is difficult to measure. Therefore, we propose to regularize the routing paths of samples according to their self-supervised similarity, \ie, considering the augmentations of an image as its similar samples. To this end, we first generate similar images in \cref{sssec:generation}. To realize consistency for similar samples and diversity for dissimilar samples, we directly regularize path distributions in \cref{sssec:consistency} and \cref{sssec:diversity}. An overview of the proposed optimization method is shown in \cref{fig.min_max}.

\subsubsection{Similar Images Generation}
\label{sssec:generation}

To obtain similar samples, we randomly augment each original sample several times by random cropping and horizontal flipping. We treat the set of augmentations as similar inputs as shown in \cref{sfg.insight_a} and \cref{sfg.insight_b}. Suppose a training batch containing $L$ samples, and $M$ augmentations for each sample. Then, we have an augmented set $\{s_{i,1}, s_{i, 2}, \cdots, s_{i,M}\}$  for a sample $s_i$, where $s_{i,j}$ stands for the $j$-th augmentation for $s_i$. Therefore, there are $L \times M$ inputs in an iteration, \ie, $\{s_{1,1}, s_{1, 2}, \cdots, s_{1,M}\}, \cdots, \{s_{L,1}, s_{L,2}, \cdots, s_{L,M}\}$.



\subsubsection{Consistency Regularization}
\label{sssec:consistency}
As shown in \cref{sfg.insight_c}, consistency regularization is an attractive force in each group of similar samples to make them closer. Let $r_{i,j}$ be the routing path of sample $s_{i,j}$, $\overline{r_{i,:}}$ be the mean of $\{r_{i,1}, r_{i,2}, \cdots, r_{i,M}\}$, which represents the routing path center of sample $s_{i}$'s augmentations.
In this way, the optimization of consistency is written as
\begin{equation}
\begin{aligned} 
\label{eq.consistent}
\mathcal{L}_{con}= & \dfrac{1}{L}\dfrac{1}{M}\sum_{i=1}^{L} \sum_{j=1}^{M}[\left \|\, r_{i,j}-\overline{r_{i,:}}\, \right \| - m_c]_+^2,
\end{aligned}
\end{equation}
where $L$ is the batch size, $M$ is the number of augmentations, 
$\| \cdot \|$ is the $L^2$ norm, $m_c$ is the margin for consistency, and $[x]_+ = max(0, x)$ denotes the hinge.

\begin{figure*}[t]
	\centering
	\includegraphics[width=\linewidth]{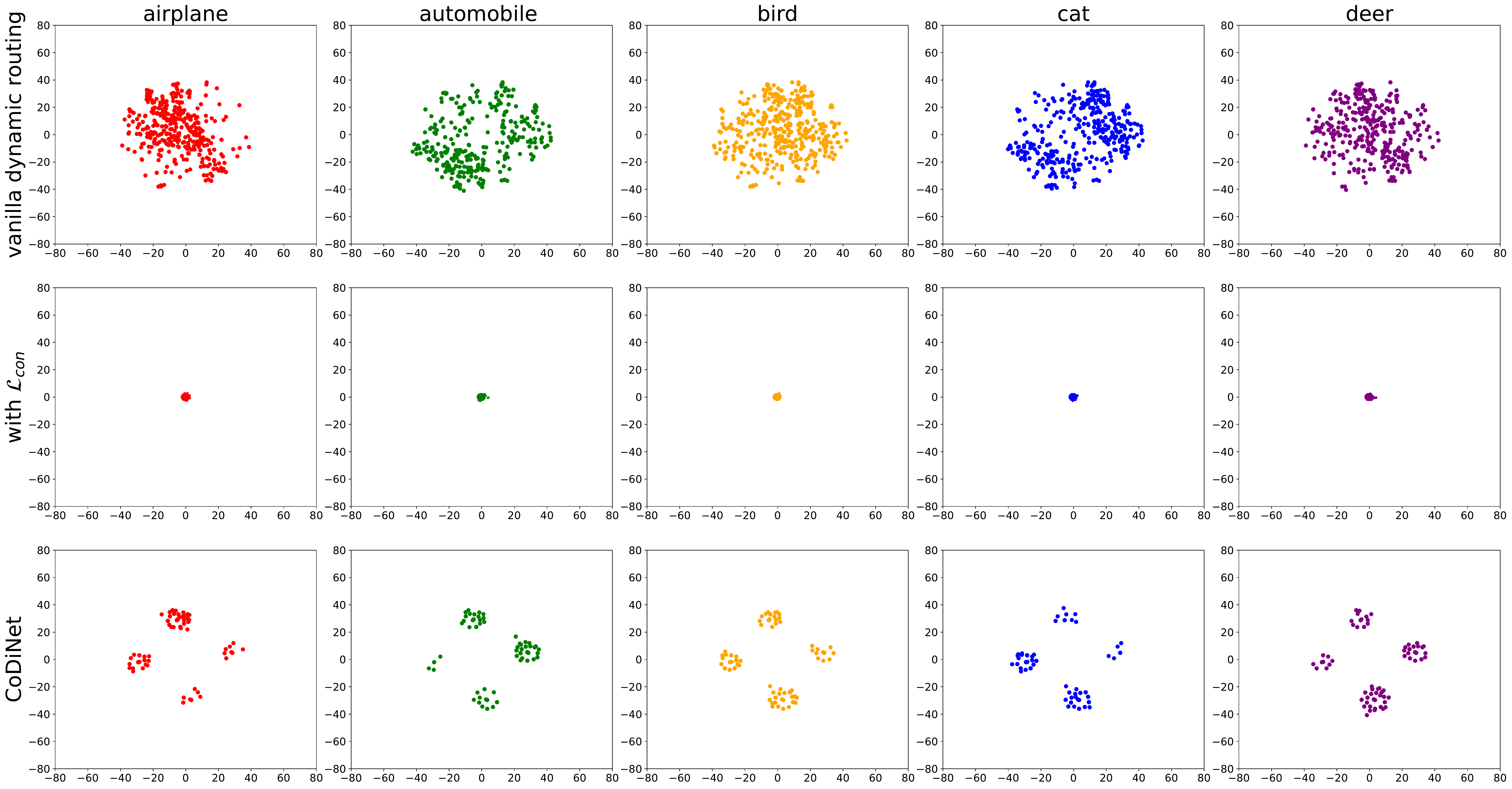}
	\caption{Visualization of the routing path distribution under different constraints through t-SNE on the CIFAR-10 test set. Figures in the first row show the path distribution of the vanilla dynamic routing method without $\mathcal{L}_{con}$ and $\mathcal{L}_{div}$. Figures in the second row show the path distribution of the experiments with $\mathcal{L}_{con}$ only. Figures in the third row show the path distribution of our method. All figures are in the same coordinate scales. 
	}
	\label{fig.consistency}
\end{figure*}

The minimization of $\mathcal{L}_{con}$ is to narrow down the differences between all the routing paths and the mean routing path. To further illustrate the consistency, we visualize the path distribution of five classes from the CIFAR-10 test set in \cref{fig.consistency}. In the first row of \cref{fig.consistency}, the paths of the vanilla dynamic routing method scatter in the routing space randomly. With consistency regularization, paths cluster around the center in the routing space, becoming consistent as shown in the second row of \cref{fig.consistency}.


\subsubsection{Diversity Regularization}
\label{sssec:diversity}
Conversely, as shown in \cref{sfg.insight_d}, diversity regularization is a repulsive force between each group of similar samples to push them away. The optimization of diversity is defined as
\begin{equation}
\label{eq.diverse}
\begin{aligned}
	\mathcal{L}_{div}= & \dfrac{1}{L}\dfrac{1}{L-1} \sum_{i=1}^{L} \sum_{j \neq i} [m_d-\left \|\overline{r_{i,:}} -\overline{r_{j,:}}\,\right \|]_+^2,
\end{aligned}
\end{equation}
where $m_d$ is the margin for diversity, and $\sum_{j\neq i}$ means all samples in the batch except $s_i$. Within a batch, the mean routing path is optimized to maximize the differences for different groups. In this way, the routing paths of different groups are dispersed and the diversity of routing paths can be guaranteed. 

Another advantage of diversity regularization is that it helps the network explore more paths. Path distribution in the real scenario under diversity regularization is also shown in \cref{fig.consistency}. Compared with the first two rows, samples in the third row cluster around several centers and different clusters keep distant from each other, which is in line with our expectations.

In our method, consistency and diversity are two facets of the problem. Although $\mathcal{L}_{con}$ makes the routing space compact, it also runs a risk of making the whole routing space collapse into a small space, impairing the diversity of routing paths. Thus, introducing $\mathcal{L}_{div}$ can compensate for the disadvantage of stand-alone $\mathcal{L}_{con}$ and make the routing paths diverse at the same time. Similarly, $\mathcal{L}_{div}$ can enhance the routing space exploration; however, making routing paths scattered without constraint can be harmful to the parameter sharing among routing paths. As a result, we propose to make use of $\mathcal{L}_{con}$ and $\mathcal{L}_{div}$ together.


\subsection{Customizable Dynamic Routing}
\label{sec.cost}

Dynamic routing is aimed at saving the cost of a network at inference time. How much cost dynamic routing should save for a network depends on the application scenario. Since the computational budget is different from device to device, 
it is better to make a dynamic routing network adaptive to devices.


We design a learning strategy to make a dynamic routing network adaptive to different computational budgets.
Let $c_k$ be the computational cost of the $k$-th block. The total cost for an $n$-block network is defined as
\begin{equation}
	{cost}_{all} = \sum_{k=1}^{n} c_k \cdot u_k.
\end{equation}

To make it learnable, we use the relaxed continuous routing variable, $v_k$, similar to \cref{eq1}. After relaxation, the loss function for cost optimization $\mathcal{L}_{cost}$ becomes
\begin{equation}
	\mathcal{L}_{cost} = \sum_{k=1}^{n} c_k \cdot v_k.
\end{equation} 
It is worth noting that we build a computational cost lookup table which records the floating-point operations (FLOPs) of each block. During optimization, each block $F_k$ will be assigned a cost $c_k$ given by the lookup table.

By putting all the losses together, the overall objective of our method is
\begin{equation}
	\mathcal{L}_{total} = \mathcal{L}_{cls} + \alpha \cdot \mathcal{L}_{con} + \beta \cdot \mathcal{L}_{div}+ \gamma \cdot \mathcal{L}_{cost},
	\label{eq.total}
\end{equation}
in which $\mathcal{L}_{cls}$ is the cross entropy loss used for classification, $\mathcal{L}_{con}$ is the loss defined in \cref{eq.consistent}, and $\mathcal{L}_{div}$ is the loss defined in \cref{eq.diverse}. $\alpha$, $\beta$, and $\gamma$ are the hyper-parameters for respective losses.

To make a dynamic routing network adaptive to different computational budgets, we tune the hyper-parameter $\gamma$ of $\mathcal{L}_{cost}$. Therefore, we can customize the learned network with different computational costs and performances. That is, a smaller $\gamma$ encourages an expensive model, and a larger $\gamma$ encourages an inexpensive model.



\begin{table*}[t]
	\caption{The Efficiency and Accuracy Trade-off based on ResNet-110 on CIFAR-10}
	\label{tab.Detailed_Data}
	\vspace{-1em}
	\renewcommand\arraystretch{1.35}
	\setlength{\tabcolsep}{4.85mm}{
		\begin{tabular}{lccccccc}
			\toprule
			\multicolumn{1}{l}{Settings}  & \multicolumn{1}{c}{GMACCs} & \begin{tabular}[c]{@{}c@{}}Speedup\\ (in GMACCs)\end{tabular} & \begin{tabular}[c]{@{}c@{}}Inference Time\\ GTX-1080 (ms)\end{tabular} & \begin{tabular}[c]{@{}c@{}}Inference Time\\ GTX-1080ti (ms)\end{tabular} & \begin{tabular}[c]{@{}c@{}}Inference Time\\ RTX-2080ti (ms)\end{tabular} & Acc. (\%)\\ \midrule
			ResNet-110  & 0.51 & 1.0$\times$ & 17.2  & 16.4  & 15.3 &  93.60 \\ \hline
			$\gamma=0.01$ & 0.29 & 1.8$\times$ & 9.43 & 8.69 & 8.56 & 94.47 \\ 
			$\gamma=0.02$ & 0.27 & 1.9$\times$ & 8.53 & 8.41 & 8.16 & 94.30 \\ 
			$\gamma=0.04$ & 0.22 & 2.3$\times$ & 7.69 & 7.62 & 7.37 & 93.94 \\ 
			$\gamma=0.08$ & 0.20 & 2.6$\times$ & 7.66 & 7.18 & 6.98 & 93.71 \\ 
			$\gamma=0.10$ & 0.10 & 5.1$\times$ & 3.85 & 3.74 & 3.29 & 92.45 \\ \bottomrule
	\end{tabular}}

	\medskip
	\emph{\footnotesize GMACCs refers to billions of multiply-accumulates. $\gamma$ is the weight for $\mathcal{L}_{cost}$ defined in Eq.(12). The unit of inference time is millisecond.}
	\vspace{-1.0em}
\end{table*}

\begin{table}[t]
	\caption{Ablation Study of Routers, $\mathcal{L}_{con}$ and $\mathcal{L}_{div}$ on CIFAR-10}
	\renewcommand\arraystretch{1.3}
	\label{tab:ablation_study}
	\vspace{-1em}
	\setlength{\tabcolsep}{2.4mm}{
		\begin{tabular}{l|ccccc}
			\toprule
			\multicolumn{1}{l|}{Methods} & Routers & $\mathcal{L}_{con}$ & \multicolumn{1}{c|}{$\mathcal{L}_{div}$} &\#Path & Acc. (\%)\\ \midrule
			ResNet-110 & --- &--- & \multicolumn{1}{c|}{---} & 1   &     93.60   \\ 
			Vanilla & $\checkmark$ &--- & \multicolumn{1}{c|}{---} & 113   &     93.66     \\ \midrule
			Vanilla + $\mathcal{L}_{con}$& $\checkmark$ &$\checkmark$ & \multicolumn{1}{c|}{---}  &20 &  		92.88   \\ 
			Vanilla + $\mathcal{L}_{div}$& $\checkmark$ & --- & \multicolumn{1}{c|}{$\checkmark$}  & 1079 &	91.34 \\  
			CoDiNet &$\checkmark$ & $\checkmark$ & \multicolumn{1}{c|}{$\checkmark$} &276 &    \textbf{94.47}     \\ \bottomrule
	\end{tabular}}

	\medskip
	\emph{\footnotesize Vanilla is the vanilla dynamic routing network. The number of activated routing paths and the accuracy under different constraints based on ResNet-110.}
\end{table}

\subsection{Training and Inference}
\label{sssec:inference}
At training time, we optimize our network in two stages. First, we train the parameters of all the blocks and all the routers together by \cref{eq.total}.
Second, we finetune our network to narrow down the  decisions' difference between training and inference. 
We use continuous decisions $(v_1, v_2, \cdots, v_n)$ as a relaxation of binary decisions $(u_1, u_2, \cdots, u_n)$ to make the first training stage end-to-end, but training with the continuous relaxation $v_k$ of the binary decision $u_k$  will inevitably cause a gap between training and inference.
Therefore, we finetune the network with the parameters in each router fixed. 

In terms of inference, a block in the network is either run or skipped exclusively. That is,
\begin{equation}
a_{k} = \begin{dcases}
F_{k}(a_{k-1}), & \text{if } v_k\geq 0.5;\\
a_{k-1}, & \text{otherwise.}
\end{dcases}
\end{equation}

\section{Experiments}
In this section, we conduct a series of experiments to evaluate the performance of CoDiNet. First, we introduce the experimental setup. Second, we perform the ablation studies for the proposed regularization. Third, we compare our results with the state-of-the-art works and other related methods. Next, we show qualitative analysis on the proposed modules in our method. In the end, we compare different routing strategies and structures. 

\subsection{Experimental Setup}
\label{detail}
\subsubsection{Datasets and Metrics}
We evaluate our method on four widely used datasets, which are CIFAR-10~\cite{krizhevsky2009learning}, CIFAR-100~\cite{krizhevsky2009learning}, SVHN~\cite{netzer2011reading}, and ImageNet~\cite{deng2009imagenet} (ILSVRC2012). CIFAR-10/100 consists of 60,000 colored images, which are resized to 32$\times$32. Out of the 60,000 images, 50,000 images are used for training, and the other 10,000 images are used for testing. SVHN includes 73,257 training images and 26,032 testing images. ImageNet contains 1,281,167 training images and 50,000 validation images that are annotated with 1,000 classes and resized to 224$\times$224. We use classification accuracy (top-1) as an evaluation metric.

\begin{table}[t]
	\caption{Results on SVHN}
	\label{tab:SVHN_result}
	\vspace{-1em}
	\setlength{\tabcolsep}{2.6mm}{
		\renewcommand\arraystretch{1.3}
		\begin{tabular}{lccc}
			\toprule
			Networks & ResNet-110 & Vanilla dynamic routing & CoDiNet  \\ \midrule
			Acc. (\%) &   94.19     &    93.15     &  94.28  \\ 
			GMACCs & 0.51 & 0.38 & 0.39	 \\ \bottomrule
	\end{tabular}}

	\medskip
	\emph{\footnotesize Results are based on ResNet-110. 
	Vanilla dynamic routing is the method without $\mathcal{L}_{con}$ and $\mathcal{L}_{div}$.}
\end{table}

\subsubsection{Implementation Details}
For data augmentation, we follow the settings as  in~\cite{veit2018convolutional, wang2018skipnet}. 
Images from CIFAR-10/100 are padded with 4 pixels on each side. Images from all datasets except SVHN are randomly cropped and horizontally flipped with a probability of 0.5. Those from SVHN are randomly cropped only.
On CIFAR-10/100 and SVHN, the lightweight ResNets~\cite{he2016deep} are adopted as the backbones, including ResNet-32, ResNet-74 and ResNet-110. On ImageNet, ResNet-50 and ResNet-101 are adopted as the backbones. Finally, the computational is measured in GMACCs, \ie, billions of multiply-accumulate operations as used in~\cite{veit2018convolutional, wang2018skipnet, wu2018blockdrop, almahairi2016dynamic}.

During training, we use SGD as an optimizer with a momentum of 0.9 and a weight decay of 1e-4. The initial learning rate is set to 0.1 and a multi-step scheduler is adopted. On CIFAR10/100, the step-wise learning rate decays by 0.1 at 150 and 200 epochs. As for ImageNet, it decays by 0.1 every 30 epochs. As for the loss hyper-parameters, $\alpha$ and $\beta$ are set to 0.2 respectively. To control the computational cost precisely, the hyper-parameter $\gamma$ for $\mathcal{L}_{cost}$ is tuned adaptive. Besides, the margin for consistency $m_c$ and the margin for diversity $m_d$ are set to 0.2 and 0.5 respectively.


\begin{table*}[t]
	\caption{Results on CIFAR-10/100}
	\label{tab:result_10_100}
	\vspace{-1em}
	\setlength{\tabcolsep}{1.7mm}{
		\renewcommand\arraystretch{1.4}
		\begin{tabular}{clccccccccc}
			\toprule
			& & \multicolumn{3}{c}{ResNet-32}      & \multicolumn{3}{c}{ResNet-74}      & \multicolumn{3}{c}{ResNet-110}     \\ \cmidrule{3-11} 
			& & Acc. (\%) & \#Param (M) & \multicolumn{1}{c|}{GMACCs} & Acc. (\%) & \#Param (M) & \multicolumn{1}{c|}{GMACCs} & Acc. (\%) & \#Param (M) & GMACCs \\ \midrule
			\multirow{3}{*}{\rotatebox{90}{CIFAR-10}} & \multicolumn{1}{l|}{ResNets}     & 92.40   &  0.46   & \multicolumn{1}{c|}{0.14}    & 93.30   &  1.13 &   \multicolumn{1}{c|}{0.34}  & 93.60 & 1.71 &   0.51  \\ 
			&\multicolumn{1}{l|}{Vanilla dynamic routing} &       91.55   &   0.49  &  \multicolumn{1}{c|}{0.09} &      93.17 &    1.21   &  \multicolumn{1}{c|}{0.18}   & 93.66   &   1.83    &   0.30   \\ 
			&\multicolumn{1}{l|}{CoDiNet} & \bf{92.48}   &    0.49   & \multicolumn{1}{c|}{0.09} &      \bf{93.61}   &    1.21   &  \multicolumn{1}{c|}{0.19}  & \bf{94.47} &  1.83 &   0.29 \\ \midrule
		
			\multirow{3}{*}{\rotatebox{90}{CIFAR-100}} & \multicolumn{1}{l|}{ResNets} &  68.7  &  0.46  &  \multicolumn{1}{c|}{0.14}  &          70.5  & 1.13  &   \multicolumn{1}{c|}{0.34}  & 71.2       &  1.71  &    0.51    \\ 
			&\multicolumn{1}{l|}{Vanilla dynamic routing} & 66.4 & 0.49  & \multicolumn{1}{c|}{0.09} & 69.7 & 1.21 & \multicolumn{1}{c|}{0.20}  & 70.3  &  1.83   &    0.24   \\ 
			&\multicolumn{1}{l|}{CoDiNet} &  \bf{69.2}		&0.49	& \multicolumn{1}{c|}{0.11} &        \bf{70.9}    & 1.21  & \multicolumn{1}{c|}{0.21} & \bf{72.9}    &   1.83   &     0.24   \\ \bottomrule
		\end{tabular}}
		
		\medskip
		\emph{\footnotesize Vanilla dynamic routing only uses our routers without $\mathcal{L}_{con}$ or $\mathcal{L}_{div}$. GMACCs refers to  billions of multiply-accumulates. \#Param is the number of parameters.}
\end{table*}

\begin{table}[t]
	\caption{Comparison with State-of-the-Arts on CIFAR-10}
	\label{tab.sota_result}
	\vspace{-1em}
	\setlength{\tabcolsep}{4.0mm}{
		\renewcommand\arraystretch{1.3}
		\begin{tabular}{llcccc}
			\toprule
			Methods     & Backbones & GMACCs & Acc. (\%) \\
			\midrule
			\emph{baseline} \\
			ResNet-32 & --- &   0.14 &  92.40  \\
			ResNet-110& --- &  	0.50 &  93.60  \\ \midrule
			\emph{dynamic routing} \\
			SkipNet~\cite{wang2018skipnet}   & ResNet-74&0.09 &  92.38 \\
			BlockDrop~\cite{wu2018blockdrop}  & ResNet-110&0.17 &  93.60 \\
			Conv-AIG~\cite{veit2018convolutional} & ResNet-110&0.41 &  94.24 \\
			IamNN~\cite{leroux2018iamnn}   & ResNet-101 & 1.10 & 94.60 \\
			CGap~\cite{du2019efficient}	  & ResNet-110 & 0.19  & 93.43\\ 
			\midrule
			\emph{early prediction}\\
			ACT~\cite{graves2016adaptive}	 & ResNet-110& 0.38 &  93.50  \\	
			SACT~\cite{figurnov2017spatially}    & ResNet-110&0.31 &  93.40 \\
			DDI~\cite{wang2020dual} & ResNet-74 &0.14 &93.88 \\
			DG-Net~\cite{shafiee2019dynamic}     & ResNet-101& 3.20 & 93.99 \\
			DG-Net (light)  & ResNet-101& 2.22 & 91.99 \\ 
			\midrule
			\emph{ours}\\
			CoDiNet-32   &ResNet-32& 0.09 &  92.48 \\ 
			CoDiNet-110   & ResNet-110&  0.29 &  \textbf{94.47} \\
			\bottomrule
	\end{tabular}}

	\medskip
	\emph{\footnotesize The results of the others are the best results reported in their papers. GMACCs refer to billions of multiply-accumulates.}
\end{table}

\subsection{Ablation Study} \label{sec_ablation}
In this part, we discuss the effectiveness of each module in CoDiNet. 
First, we perform the ablation studies on consistency and diversity. Then, we discuss the customizable dynamic routing module, which is proposed to strike the balance between computational cost reduction and accuracy.

\begin{table}[t]
	\caption{Comparison with State-of-the-Arts on ImageNet}
	\label{tab.sota_imagenet}
	\vspace{-1em}
	\renewcommand\arraystretch{1.3}
	\setlength{\tabcolsep}{3.4mm}{
		\begin{tabular}{llcc}
			\toprule
			Methods   &  Backbones & GMACCs & Acc. (\%) \\ \midrule
			\emph{baseline} & & &\\
			ResNet-50  &  --- & 3.86 & 75.36 \\ 
			ResNet-101  &  ---  & 7.63 & 76.45 \\ 
			\midrule
			\emph{dynamic routing} & & &\\
			Conv-AIG 50~\cite{veit2018convolutional} & ResNet-50  & 3.06 & 76.18 \\
			Conv-AIG 101 & ResNet-101  & 5.11 & 77.37\\
			SkipNet~\cite{wang2018skipnet}  & ResNet-101 & 6.70 & 77.40 \\
			SkipNet (light) & ResNet-101 & 3.60 & 75.22 \\
			LC-Net~\cite{xia2020fully} & ResNet-50  & 2.89 & 74.10 \\
			BlockDrop~\cite{wu2018blockdrop} & ResNet-101  & 7.32 & 76.80 \\
			DG-Net~\cite{shafiee2019dynamic} & ResNet-101  & 7.05 & 76.80\\ 
			DDI~\cite{wang2020dual} & DenseNet-201 & 3.50 & 76.50 \\ \midrule
			\emph{early prediction} & & &\\
			MSDN~\cite{Huang2017MultiScaleDN} & DenseNets  & 2.30 & 74.24 \\
			RA-Net~\cite{yang2020resolution} & DenseNets  & 2.40 & 75.10 \\ 
			IamNN~\cite{leroux2018iamnn}  &  ResNet-101  & 4.00 & 69.50 \\ 
			ACT~\cite{graves2016adaptive} & ResNets  & 6.70  & 75.30 \\
			SACT~\cite{figurnov2017spatially}    & ResNets & 7.20 & 75.80 \\
			\midrule
			\emph{ours} & & &\\
			CoDiNet-50 & ResNet-50& 3.10 & 76.63 \\
			CoDiNet-101   & ResNet-101& 5.02  & \textbf{77.85}\\ \bottomrule
	\end{tabular}}

	\medskip
	\emph{\footnotesize The results of the others are the best results reported in their papers. GMACCs refer to billions of multiply-accumulates.}
\end{table}

\subsubsection{Effectiveness of Regularization}
First, we conduct experiments on CIFAR-10 to show the ablation studies on each component based on ResNet-110. As shown in \cref{tab:ablation_study}, the accuracy of the vanilla dynamic routing method is 93.66\% with 113 paths. With consistency regularization, limited paths are utilized, and the performance decreases to 92.88\%. When utilizing the consistency and diversity regularization at the same time, it achieves an accuracy of 94.47\% with 276 paths.

We also provide the numerical improvement on other datasets to verify the proposed consistency and diversity-based optimization as a whole. As shown in \cref{tab.sota_imagenet}, our method achieves a comparable result with 18.4\% computational cost reduction on ImageNet with 1.2\% accuracy improvements. 
The results on SVHN are shown in \cref{tab:SVHN_result}. Compared with the result of the vanilla dynamic routing method on SVHN, our method gains 1.13\% improvement with about 5\% extra computational cost, which is about 22\% computation reduction against original ResNet-110. 

\begin{figure*}[t]
	\centering
	\includegraphics[width=\linewidth]{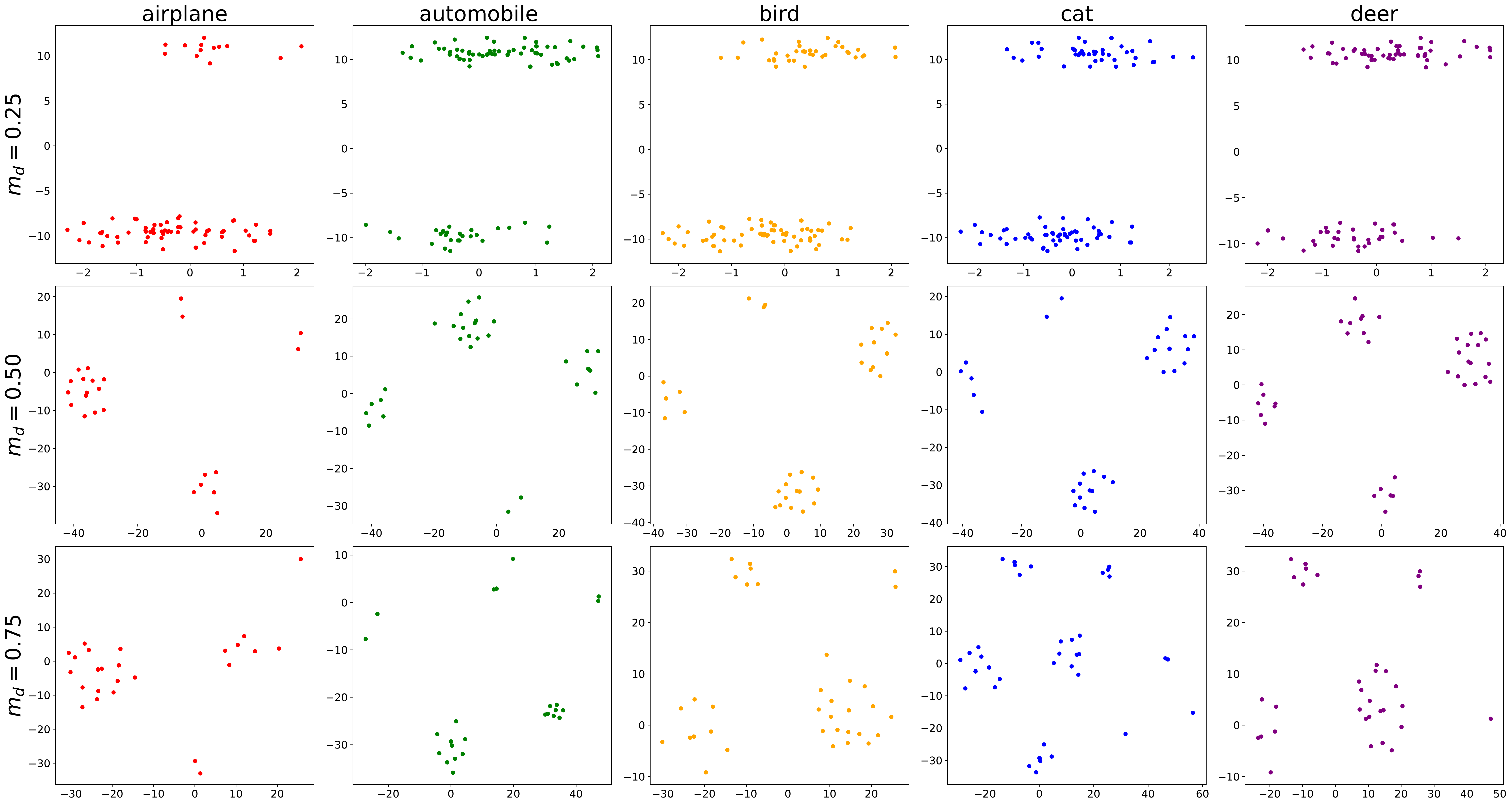}
	\caption{Visualization of the routing paths distribution under different $m_d$ through t-SNE. $m_d$ is the margin of diversity, which is defined in \cref{eq.diverse}. Figures in row 1, 2 and 3 show routing path distributions for $m_d=0.25, 0.5$, and $0.75$ respectively. Images are of 5 classes (airplane, automobile, bird, cat, and deer) from the CIFAR-10 test set.}
	\label{fig.diversity}
\end{figure*}

\begin{figure}[t]
	\centering
	\subfigure[CIFAR-10]{
		\centering
		\includegraphics[width=.485\linewidth]{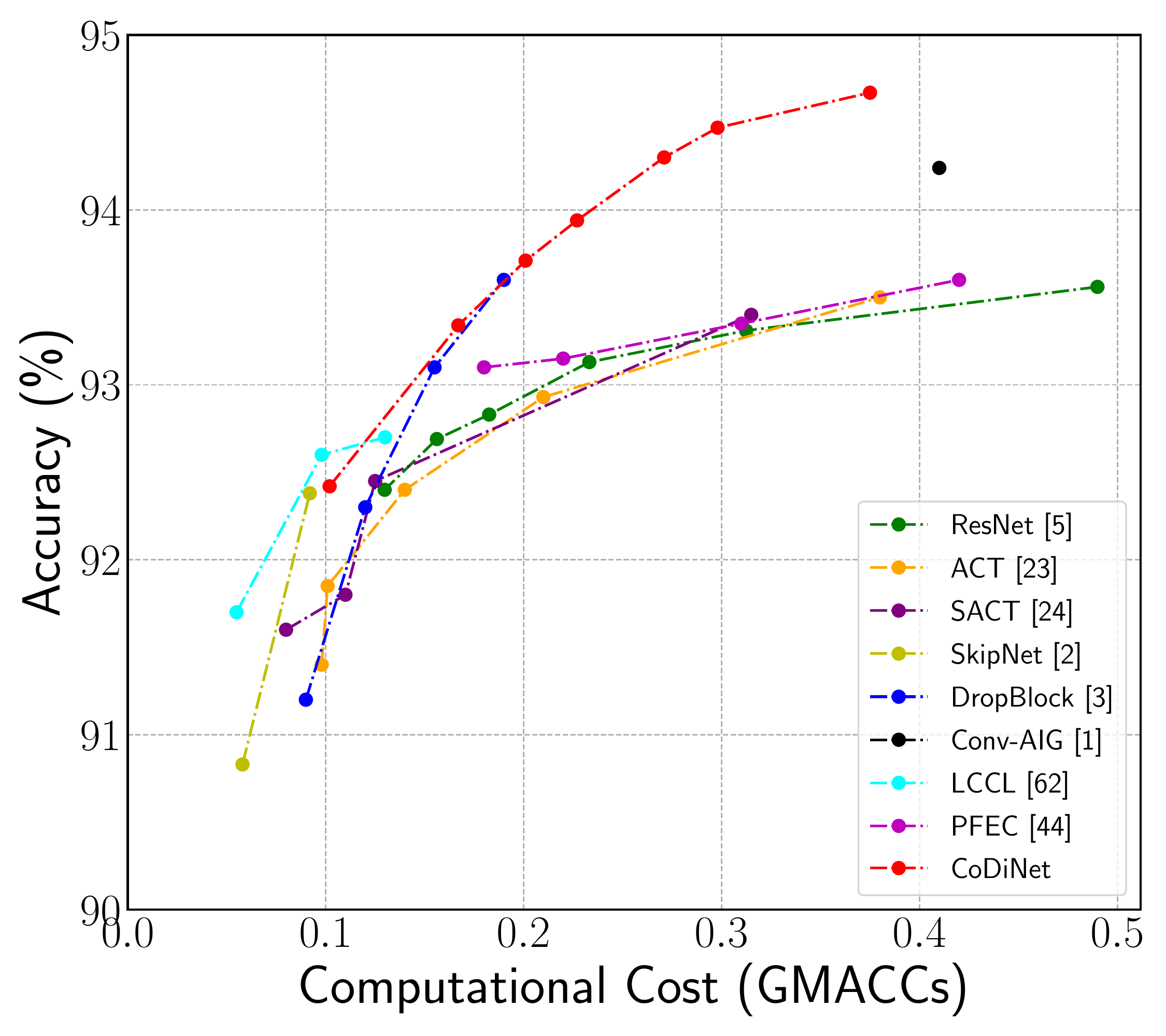}
		\label{sfg.flop_compare_a}
	}
	\hspace{-1.4em}
	\subfigure[ImageNet]{
		\centering
		\includegraphics[width=.485\linewidth]{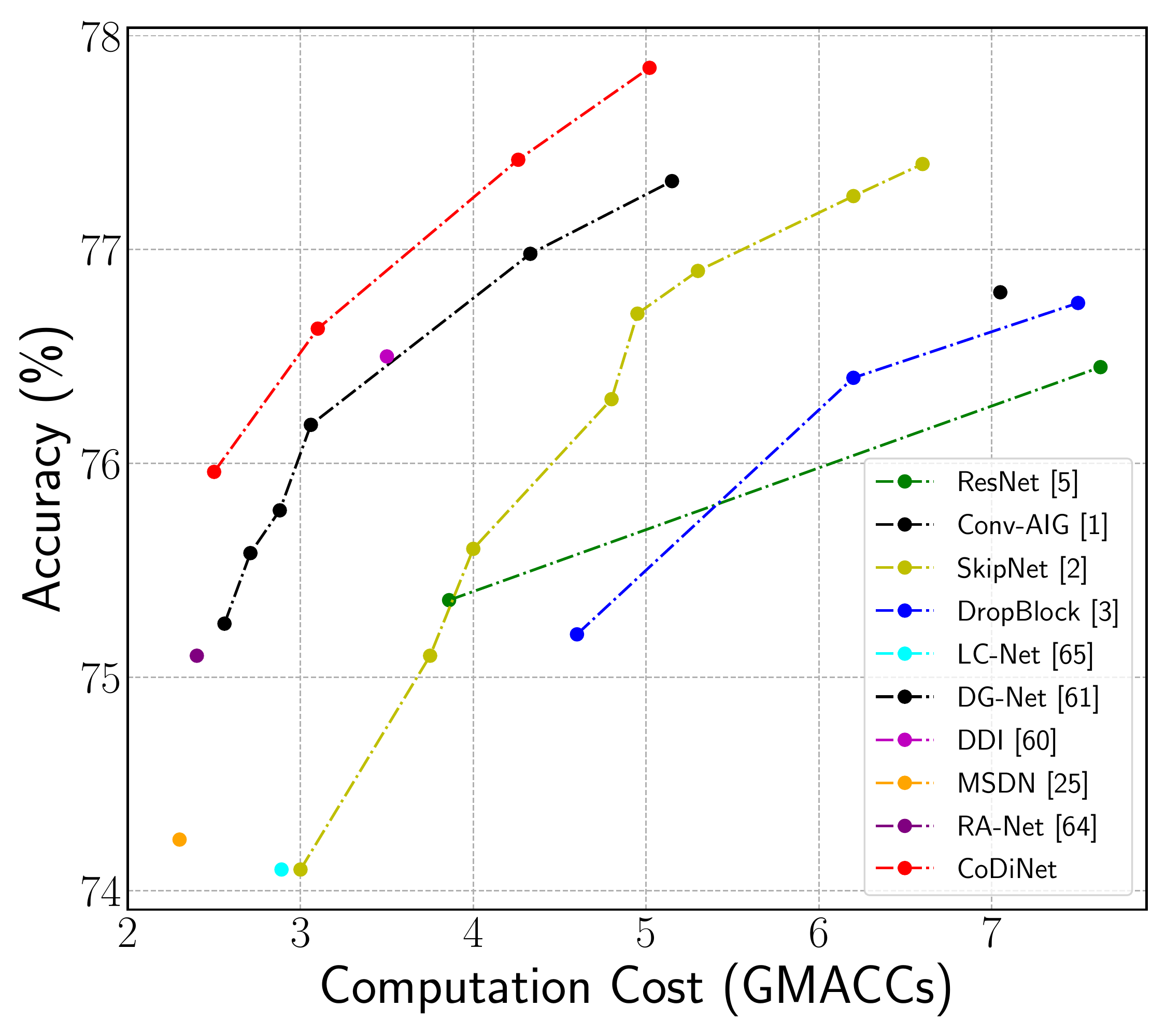}
		\label{sfg.flop_compare_b}
	}
	\caption{The accuracy against computational cost (GMACCs) of CoDiNet comparing to related methods on CIFAR-10 and ImageNet.}
	\label{fig.results_comparison}
\end{figure}

\subsubsection{Effect of Customizable Dynamic Routing}
Another benefit of our method is that we can optimize the computational cost explicitly. As we discussed in \cref{sec.cost}, the proposed $\mathcal{L}_{cost}$ can balance the trade-off between accuracy and cost. And we only need to tune the weight of $\mathcal{L}_{cost}$, \ie, $\gamma$ in \cref{eq.total}, to obtain a desired model. \cref{tab.Detailed_Data} shows the trade-off between classification accuracy, GMACCs, and the average inference time on the CIFAR-10 dataset. With the increasing of the computational cost, the accuracy tends to be upward. In the extreme case, our method achieves even only 0.10 GMACCs with a comparable accuracy, which can meet the requirements on low power platforms. 

\subsection{Performance Comparison}
In this part, we compare the results of CoDiNet with related methods. First, we compare CoDiNet with the results of ResNets on CIFAR-10/100. Next, we compare CoDiNet with the state-of-the-art dynamic routing networks, early prediction models\footnote{The results of ACT and SACT are quoted from~\cite{wu2018blockdrop} because ACT and SACT did not report corresponding results.}, and related compression methods on CIFAR-10 and ImageNet.

\subsubsection{Comparison with ResNets}
We make a comparison between CoDiNet and ResNets, w.r.t., accuracy and GMACCs. As shown in \mbox{\cref{tab:result_10_100}}, our method achieves higher accuracy with less cost in all experimental settings. In particular, compared with ResNet-110 on the CIFAR-10 dataset, our method needs 60\% cost (0.29 GMACCs) compared to the original network (0.51 GMACCs), and achieves 0.87\% improvement on accuracy. Similarly, on CIFAR-100, our method achieves significant improvement compared with ResNets and vanilla dynamic routing networks (without $\mathcal{L}_{con}$ and $\mathcal{L}_{div}$). It achieves an accuracy of 72.9\% with 0.24 GMACCs. Besides, the cost reduction on deep networks is much larger than cost reduction on shallow networks. It shows that deep networks are more redundant than shallow networks.

\subsubsection{Comparison on CIFAR-10}
\label{sssec:sotas}

We compare CoDiNet with other state-of-the-art dynamic routing methods, early prediction networks, and related compression methods. 
As shown in \cref{tab.sota_result}, we compare with the following methods: BlockDrop~\cite{wu2018blockdrop}, SkipNet~\cite{wang2018skipnet}, Conv-AIG~\cite{veit2018convolutional}, ACT~\cite{graves2016adaptive}, SACT~\cite{figurnov2017spatially}, CGAP~\cite{du2019efficient}, DDI~\cite{wang2020dual}, Iamm~\cite{leroux2018iamnn}, DG-Net~\cite{shafiee2019dynamic}. Following \cite{wu2018blockdrop}, PFEC \cite{li2016pruning} and LCCL \cite{dong2017more} are used for comparison. BlockDrop and SkipNet are prevalent methods, applying reinforcement learning and LSTM respectively to implement the dynamic routing. BlockDrop achieves an accuracy of $93.6\%$ with 0.17 GMACCs on CIFAR-10 with ResNet-110. Conversely, SkipNet focuses more on computational cost reduction, obtaining an accuracy of $92.38\%$ with 0.09 GMACCs. Conv-AIG achieves an accuracy of 94.24\% with about 0.41 GMACCs. 
As shown in \cref{sfg.flop_compare_a}, the CoDiNet outperforms other methods in most cases with a comparable computational cost. Our method achieves an accuracy of 94.47\% with only 0.29 GMACCs. More importantly, our method does not conflict with compression methods, and it can be used along with compression methods for better performance.



\subsubsection{Comparison on ImageNet}
We compare CoDiNet with state-of-the-art methods on ImageNet as shown in \cref{tab.sota_imagenet}, and the efficiency-accuracy trade-off in \cref{sfg.flop_compare_b}. Among these methods, Conv-AIG~\cite{veit2018convolutional} reports results based on ResNet-50 and ResNet-101, which are 76.18\% and 77.37\% with 3.06 and 5.11 GMACCs respectively. SkipNet~\cite{wang2018skipnet} achieves an accuracy of 75.22\% with 3.6 GMACCs. Besides, RA-Net~\cite{yang2020resolution} is an early prediction method that processes different samples in different resolutions, which achieves an accuracy of 75.10\% with 2.40 GMACCs. In comparison, CoDiNet outperforms these methods, which achieves an accuracy of 76.63\% with 3.10 GMACCs based on ResNet-50 and an accuracy of 77.85\% with 5.02 GMACCs based on ResNet-101.


\begin{figure}[t]
	\centering
	\includegraphics[width=\linewidth]{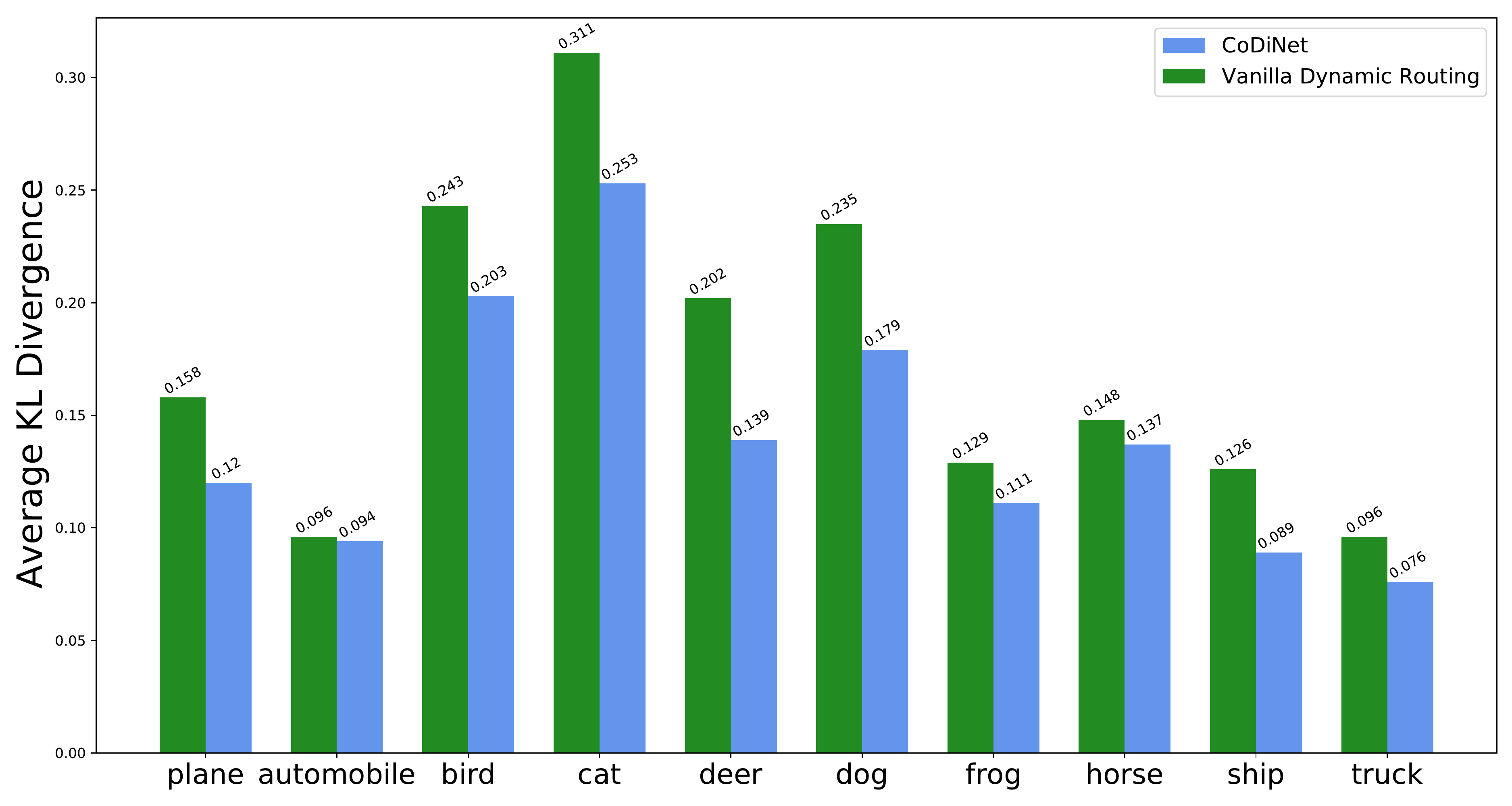}
	\caption{KL divergence of the predictions between original and its augmentation on the CIFAR-10 test set. Green bars stands for the vanilla dynamic routing ResNet-110. Blue bars stands for the dynamic ResNet-110 with $\mathcal{L}_{con}$. Best viewed in color.}
	\label{fig.KD}
\end{figure}

\begin{figure}[t]
	\centering
	\includegraphics[width=\linewidth]{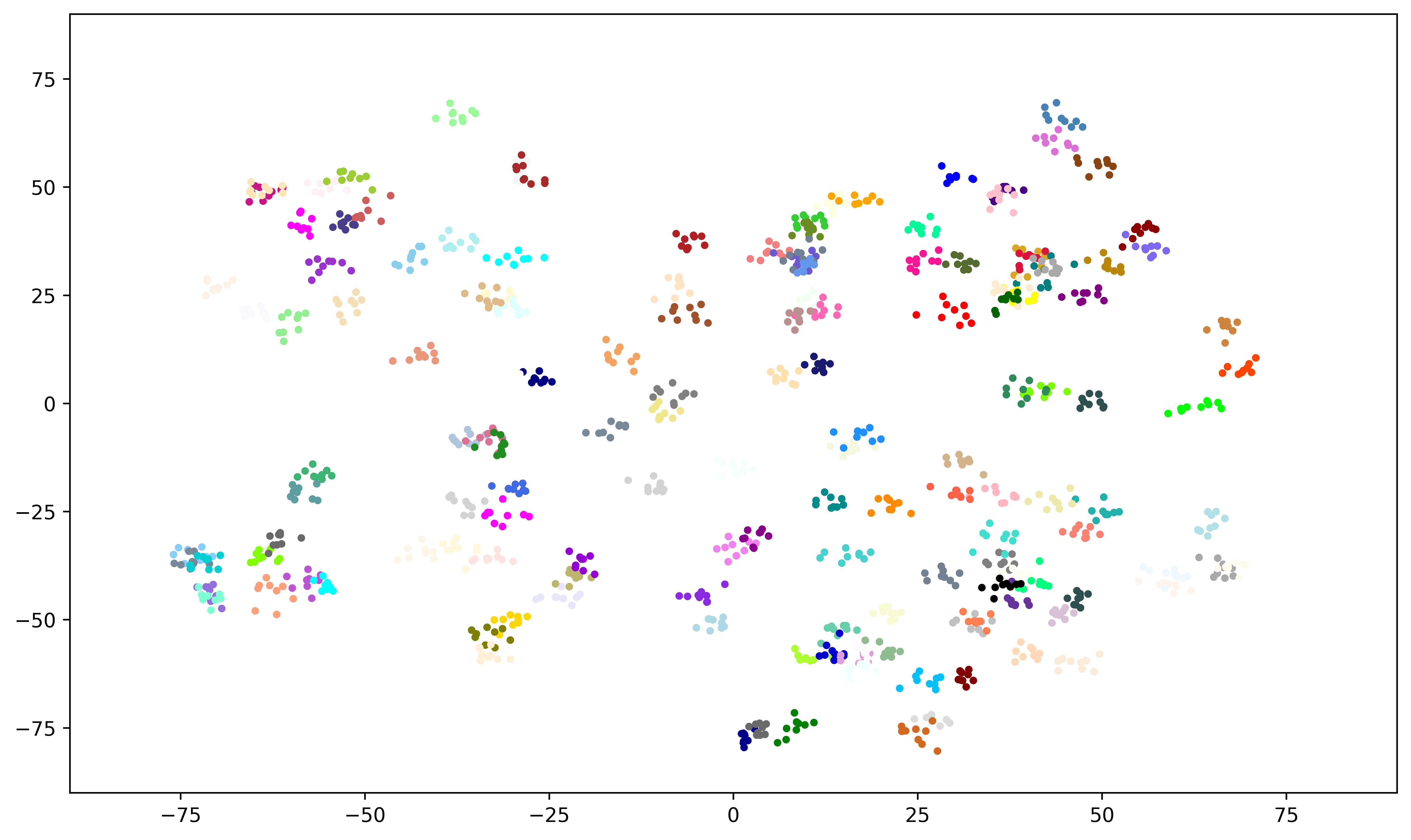}
	\caption{The visualization of the routing paths of multi-augmentations for samples. We randomly visualize 150 groups of augmentations from the CIFAR-10 test set. Dots in the same color stand for a group of augmentations from the same sample. Best viewed in color.}
	\label{fig.std_vis}
\end{figure}
\subsection{Qualitative Analysis}

In this part, we conduct experiments to qualitatively analyze our proposals. First, we analyze the effect of the consistency regularization and the diversity regularization. Next, we visualize the distribution of relaxed routing paths. Finally, we show the images sharing the same routing paths.

\begin{table}[t]
	\caption{Analysis of the Margin of Diversity}
	\renewcommand\arraystretch{1.3}
	\label{tab:path_number}
	\vspace{-1.0em}
	\setlength{\tabcolsep}{6.8mm}{
		\begin{tabular}{c|ccc}
			\toprule
			$m_d$ &\#Path & GMACCs & Acc. (\%) 		\\ \midrule
			0.25 &196     & 0.22  &   93.12       \\ 
			0.50 &276     & 0.29  &   \textbf{94.47} \\ 
			0.75 &396     & 0.24 &    92.75      \\ 
			1.00 &800     & 0.32 &    92.46      \\ \bottomrule
	\end{tabular}}

	\medskip
	\emph{\footnotesize The number of activated routing paths and accuracy under different margins of diversity with ResNet-110 on CIFAR-10. $m_d$ is the margin of diversity defined in \cref{eq.diverse}, \#Path is the number of utilized routing paths and GMACCs refer to billions of multiply-accumulates.}
\end{table}


\subsubsection{Analysis of Consistency} 
\label{sssec:ana_consistency}
To illustrate the effectiveness of the consistency regularization, we adopt KL divergence as an indicator to measure the difference of paths between original test set and augmented test set. That is, a smaller KL divergence indicates a better consistency. Compared with the vanilla dynamic routing on the CIFAR-10 test set, we found that the consistency regularization can significantly enhance the consistency between routing paths of original and augmented images. It also improves the performance of the augmented test set. As shown in \cref{fig.KD}, the KL divergence between original and augmented images decreases considerably with $\mathcal{L}_{con}$.


Additionally, we design a qualitative experiment to show the effect of the consistency regularization. We adopt various augmentation methods including random cropping, horizontal flipping, vertical flipping, and rotation on the 10,000 images of the CIFAR-10 test set. For each original image, we compare its routing path with that of its augmentation 
under two models: CoDiNet and vanilla dynamic routing. As a result, 5,342 out of 10,000 image pairs have consistent routing paths with CoDiNet, \ie, the original image and its augmented image have the same routing path. With vanilla dynamic routing, only 2,265 images have consistent paths with their augmentation. 

\subsubsection{Analysis of Diversity}
\label{sssec:ana_diversity}

The number of distinct routing paths at inference time under different settings of $m_d$ is shown in \cref{tab:path_number}. With a larger margin, more routing paths will be obtained. When $m_d$ is 0.5, our method achieves the best performance on CIFAR-10 based on ResNet-110. When $m_d$ is larger than 0.5, the performance drops. The reason for that might lie in too dispersed routing paths resulting in under-fitting.

Next, we visualize the routing path distribution on the CIFAR-10 test set to demonstrate the tendency of different margins of diversity $m_d$. In \cref{fig.diversity}, paths are dispersed significantly with an increase of $m_d$. When $m_d$ is 0.25, paths cluster into two groups. When $m_d$ is 0.75, the scale of coordinate is similar as $m_d = 0.5$, but the distribution is more disperse. 

\begin{figure*}[t]
	\centering
	\subfigure[Path distribution without $\mathcal{L}_{con}$ or $\mathcal{L}_{div}$]{
		\centering
		\includegraphics[width=0.48\linewidth, height=2.3in]{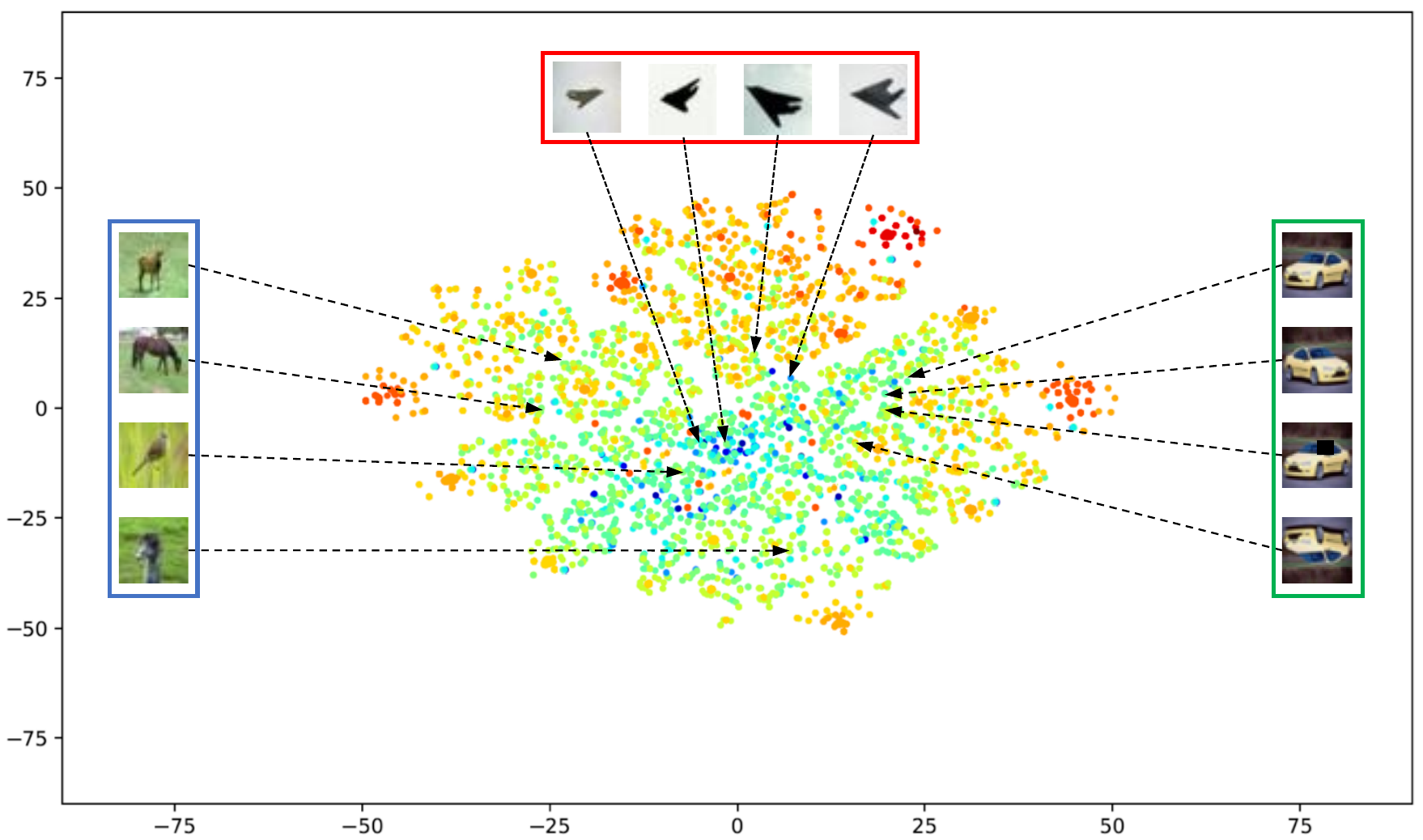}
		\label{sfg.disconsist}
	}
	\subfigure[Path distribution with $\mathcal{L}_{con}$ and $\mathcal{L}_{div}$]{
		\centering
		\includegraphics[width=0.48\linewidth, height=2.3in]{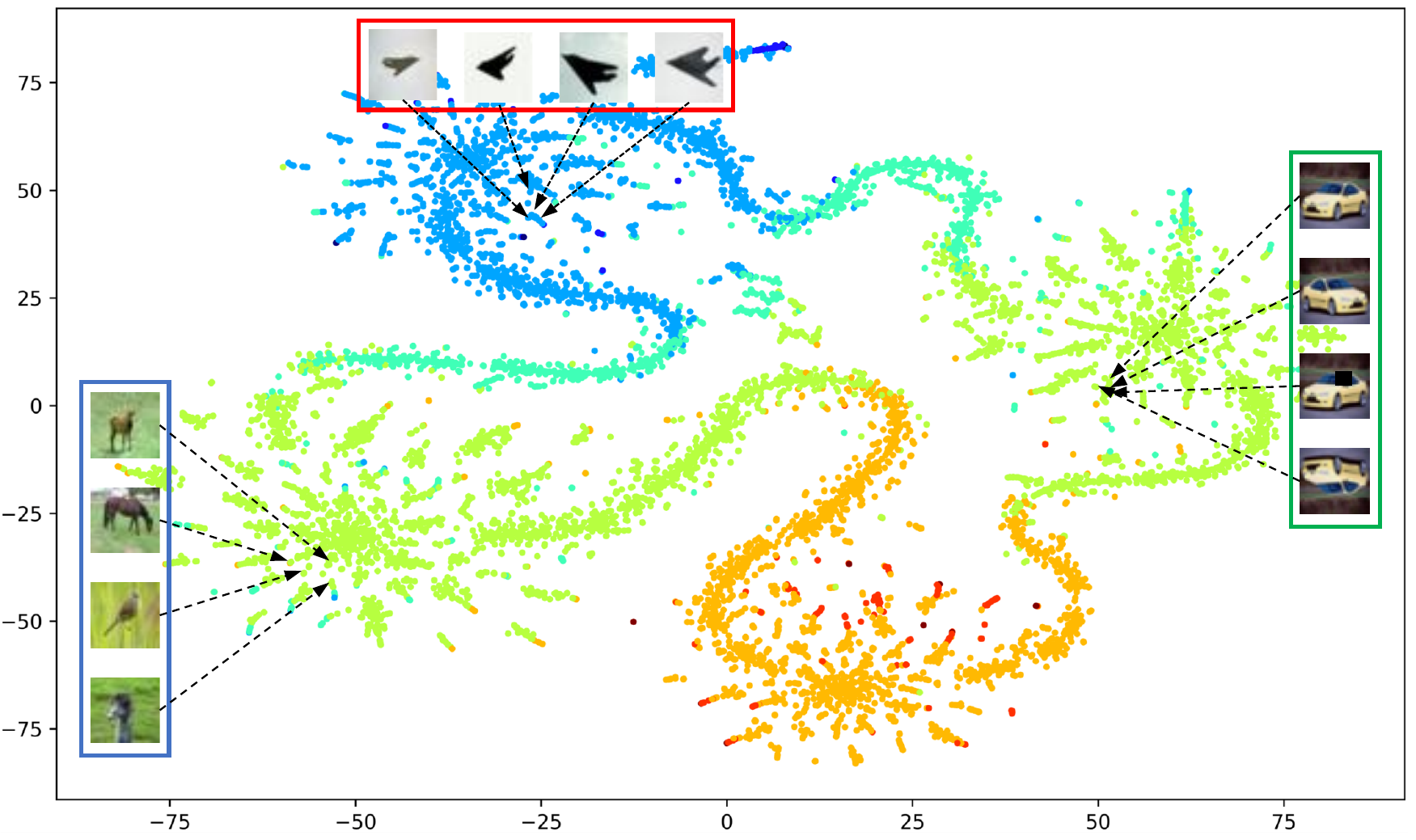}
		\label{sfg.consist}
	}
	\caption{Visualization of continuous routing paths by t-SNE on the CIFAR-10 test set. 
	The color of each dot is the mathematical expectation numbers of to-be-run blocks. Red refers to more to-be-run blocks, while blue to fewer. Images in the green rectangle are augmentations from the same image. Images in the red rectangle are from the same category. Images in the blue rectangle are from different categories. Best viewed in color.}
	\label{fig.similar}
\end{figure*}
\subsubsection{Illustration of Routing Paths for Similar Samples}

In this section, we obtain groups of similar images by applying self-supervised augmentation methods (random cropping, horizontal flipping, vertical flipping, and rotation) on the CIFAR-10 test set ten times and then visualize the distribution of routing paths for self-supervised similar images to show the routing paths of ``similar samples'' clustering together by directly visualizing the distribution of routing paths for self-supervised similar images. We mark the augmented images from the same raw images with the same color. As shown in \cref{fig.std_vis}, points with the same color, \ie, augmentations of the same sample, tend to cluster together. As a result, the routing paths of self-supervised similar samples tend to cluster together in our method.

\subsubsection{Correlation between Samples Similarity and Routing Paths Similarity} 

\begin{figure}[t]
    \centering
	\subfigure[Vanilla dynamic routing]{
		\includegraphics[width=0.485\linewidth]{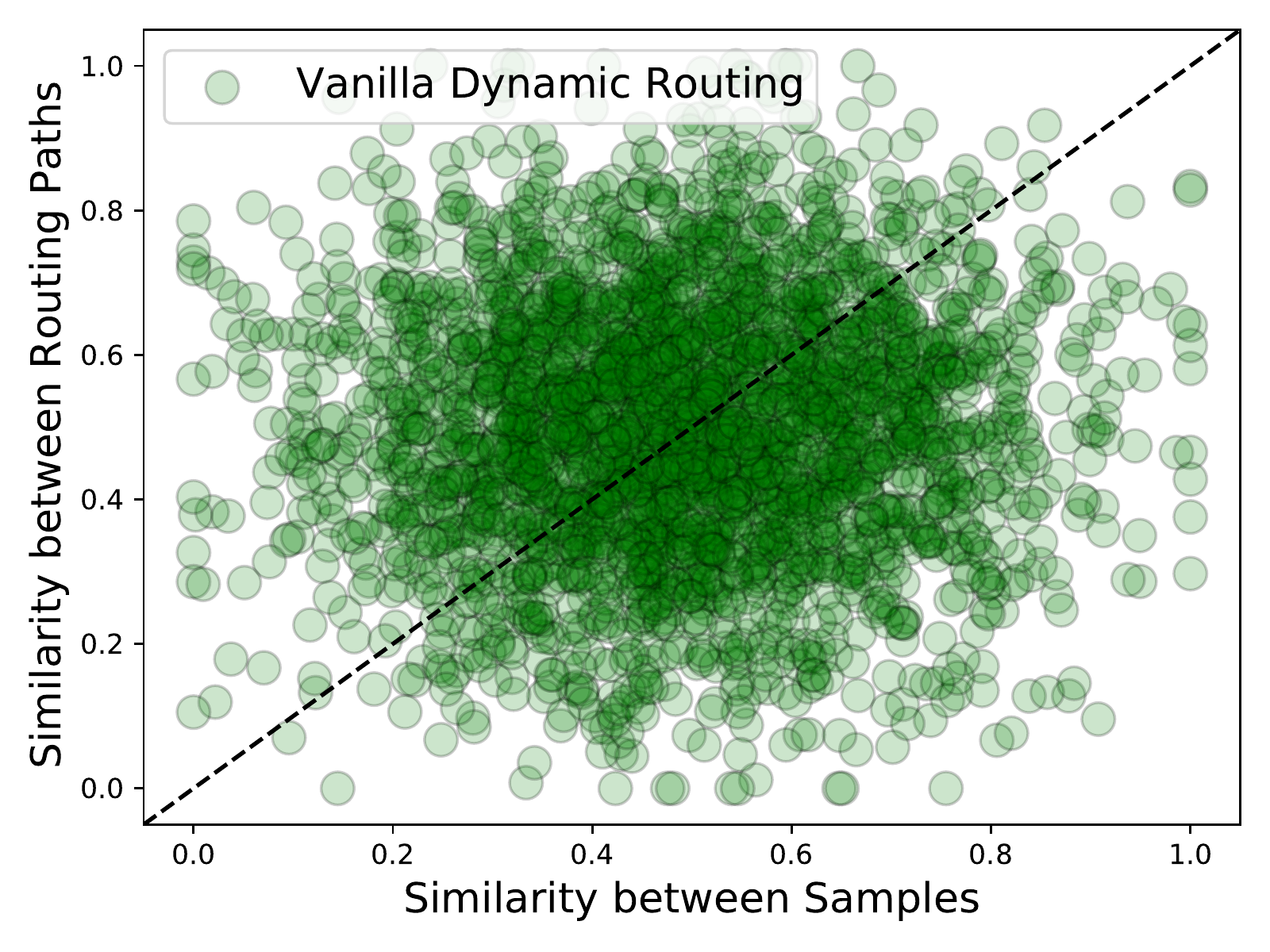}
		\label{sfg.NMI}
	}
	\hspace{-1.4em}
	\subfigure[Our CoDiNet]{
		\includegraphics[width=0.485\linewidth]{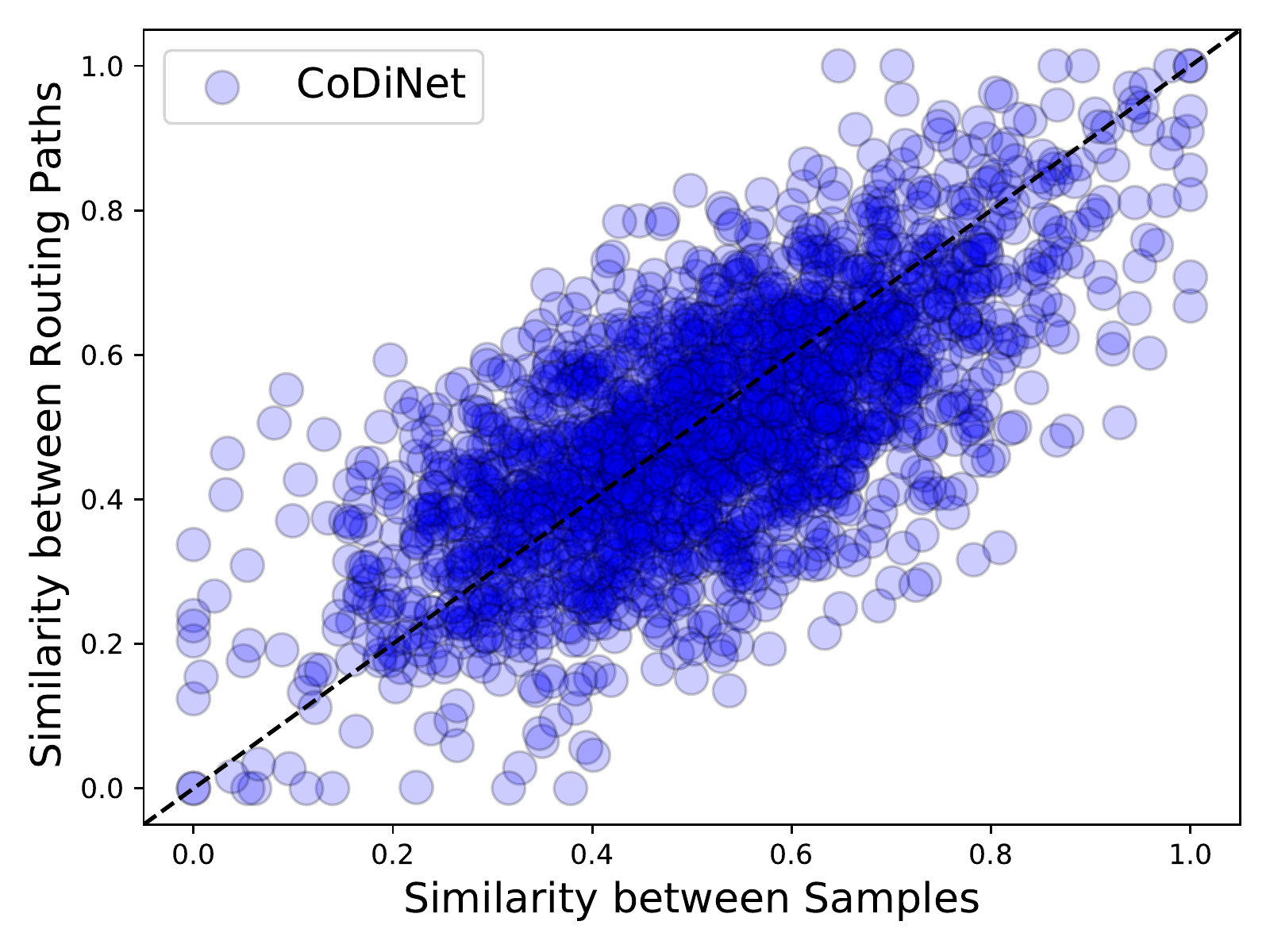}
		\label{sfg.EUR}
	}
    \caption{The correlation between the similarity of sample features and the similarity of sample routing paths with our CoDiNet and the vanilla dynamic routing. We use Cosine Similarity to measure the similarity. It is worth noting that the closer the point to the dotted line, the more positively correlated the similarity of samples and the similarity of paths.}
    \label{fig.similar_core}
\end{figure}

In this section, we calculate the PCC (Pearson correlation coefficient) between sample feature similarity and sample routing path similarity. A higher correlation coefficient value indicates the sample feature similarity is more positively correlated to the path similarity, therefore, the routing paths of similar images are closer in the routing space.  
Specifically, we use the Cosine Similarity on every pair of routing paths and sample features as the routing paths similarity and sample similarity. To better represent samples, the sample features are extracted by a third-party unsupervised model (an ImageNet pre-trained MoCo~\cite{moco}). As shown in \cref{tab.similarity_method}, the Pearson correlation coefficient for our CoDiNet is 0.581, while the one for the vanilla dynamic routing is 0.024. Thus, our method is about 24 times larger than the vanilla one. Besides, we plot the correlation diagrams for different sample pairs in \cref{fig.similar_core}. Clearly, our method is more likely to encourage the consistency between routing path similarity and sample similarity.


\subsubsection{Visualization of Relaxed Routing Paths} 

As shown in \cref{fig.similar}, we visualize the relaxed routing paths of the vanilla dynamic routing network and CoDiNet by t-SNE. Different colors correspond to different mathematical expectations of the numbers of to-be-run blocks. Red refers to more to-be-run blocks, while blue refers to less to-be-run block. The path distribution of the method without $\mathcal{L}_{con}$ or $\mathcal{L}_{div}$, is shown in \cref{sfg.disconsist}, where the paths gather in a small space around the center. In comparison, the path distribution of CoDiNet are regularly distributed and scattered throughout a much larger space as shown in \cref{sfg.consist}.


\begin{table}[t]
	\caption{Illustration of the PCC (Pearon correlation coefficient) between sample similarity and the routing path similarity.}
	\vspace{-1.0em}
	\label{tab.similarity_method}
	\setlength{\tabcolsep}{4.4mm}{
		\begin{tabular}{lcc}
			\toprule
			Methods & Vanilla Dynamic Routing & Our CoDiNet \\ \midrule
			PCC & 0.024 & 0.581 \\ \bottomrule
        \end{tabular}}

	\medskip
	\emph{\footnotesize PCC refers to the Pearon correlation coefficient. The PCC value ranges from -1 to 1. A PCC value of 0 implies that there is no linear correlation between the similarities. Experiments is on CIFAR-10.}
\end{table}

Moreover, we provide three groups of samples to present the routing paths of the self-supervised similar images cluster together no matter they belong to the same category or not. As shown in~\cref{fig.similar}, we provide three groups of images and mark out their routing paths. Firstly, images in the green rectangle are augmentations from the same image. Then, images in the red rectangle are from the same category. Next, images in the blue rectangle are from different categories. As shown in~\cref{sfg.disconsist}, the routing paths of images in each rectangle are scattered among the whole distribution without $\mathcal{L}_{con}$ or $\mathcal{L}_{div}$. In comparison, with $\mathcal{L}_{con}$ and $\mathcal{L}_{div}$, the routing paths of images in each group are respectively cluster together as shown in~\cref{sfg.consist}. Therefore, the proposed consistency regularization term can effectively make the routing paths of similar images clustering together.

\section{Conclusion}
In this paper, we see routing mechanisms from a novel perspective that regards a dynamic routing network as a mapping from a sample space to a routing space. From this view, path distribution in routing space is a fundamental problem in a dynamic routing network.  We propose a novel framework CoDiNet to regularize path distribution with diversity and consistency. Moreover, we design a customizable dynamic routing module enabling the network to adapt to different computational budgets. 
We compare CoDiNet with state-of-the-art methods on four benchmark datasets, demonstrating that it can effectively reduce the computational cost without compromising performance. 

\section*{Acknowledgement}
This work is supported in part by National Key Research and Development Program of China under Grant 2020AAA0107400, National Natural Science Foundation of China under Grant U20A20222, Zhejiang Provincial Natural Science Foundation of China under Grant LR19F020004, and key scientific technological innovation research project by Ministry of Education.

\bibliographystyle{IEEEtran}
\bibliography{bib}

\newcounter{mybibstartvalue}
\setcounter{mybibstartvalue}{62}

\xpatchcmd{\thebibliography}{%
  \usecounter{enumiv}%
}{%
  \usecounter{enumiv}%
  \setcounter{enumiv}{\value{mybibstartvalue}}%
}{}{}

\renewcommand{\refname}{\vspace{-3em}}


\begin{IEEEbiography}[{\includegraphics[width=1in,height=1.25in,clip,keepaspectratio]{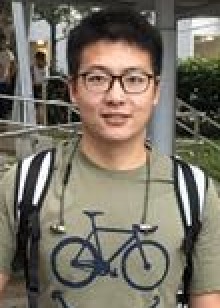}}]{Huanyu Wang}
  received his master's degree in 2017 from National University of Singapore, Singapore, where he worked on problems in machine learning and massive data. He is currently a Ph.D. candidate at Zhejiang University. His current research interests include dynamic routing, semantic segmentation, and neural architecture search.
\end{IEEEbiography}

\begin{IEEEbiography}[{\includegraphics[width=1in,height=1.25in,clip,keepaspectratio]{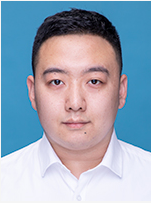}}]{Zequn Qin}
  received his master's degree in 2019 from Northwestern Polytechnical University, China, where he worked on problems in computer vision and pattern recognition. He is currently a Ph.D. candidate at Zhejiang University. His current research interests include autonomous vehicles, dynamic routing, and semantic segmentation.
\end{IEEEbiography}

\begin{IEEEbiography}[{\includegraphics[width=1in,height=1.25in,clip,keepaspectratio]{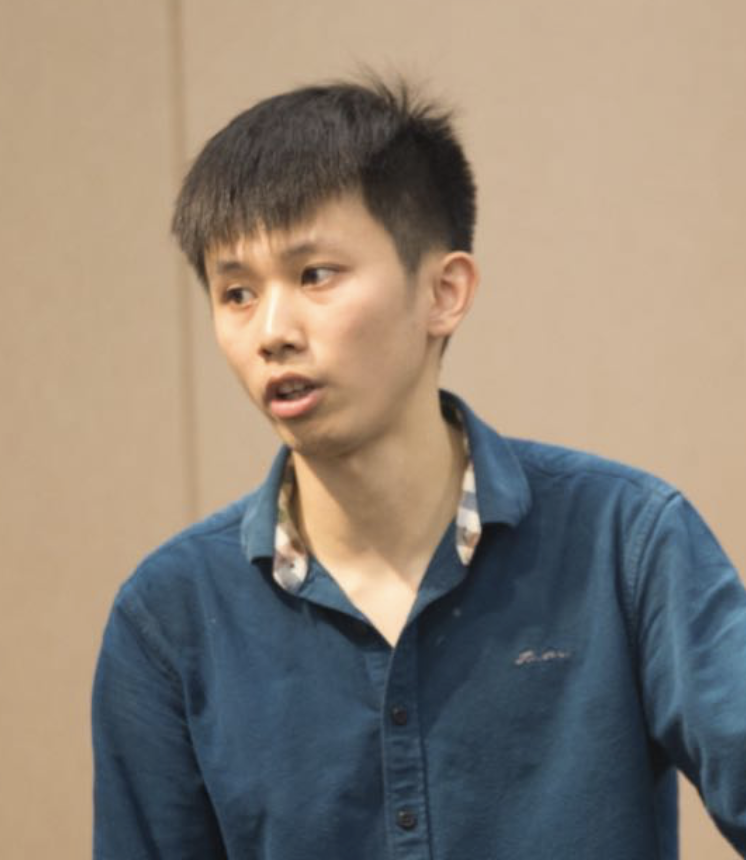}}]{Songyuan Li}
  received his master's degree in 2017 from Zhejiang University, China, where he worked on problems in computer architecture and operating systems. He is currently a Ph.D. candidate at Zhejiang University. His current research interests include semantic segmentation and dynamic routing.
\end{IEEEbiography}

\begin{IEEEbiography}[{\includegraphics[width=1in,height=1.25in,clip,keepaspectratio]{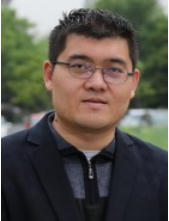}}]{Xi $\textup{Li}^{\dagger}$}
  received his Ph.D. degree in 2009 from the National Laboratory of Pattern Recognition, Chinese Academy of Sciences, Beijing, China. From 2009 to 2010, he was a Post-Doctoral Researcher with CNRS Telecom ParisTech, France. He was a Senior Researcher with the University of Adelaide, Australia. He is currently a Full Professor with Zhejiang University, China. His research interests include visual tracking, compact learning, motion analysis, face recognition, data mining, and image retrieval.
\end{IEEEbiography}

\newpage
\appendices
\section{Discussion on Routers}




In this section, we compare CoDiNet with two implementations of Gumbel-Softmax, and different router structures to analyze the design of our router. The discussion focuses on the following questions. First, which variant of Gumbel-Softmax is suitable to utilize in our method? Next, what are the advantages of the router used in our method compared to other kinds of routers? 

\begin{figure}[h]
	\centering
	\subfigure[CIFAR-10]{
		\centering
		\includegraphics[width=.487\linewidth]{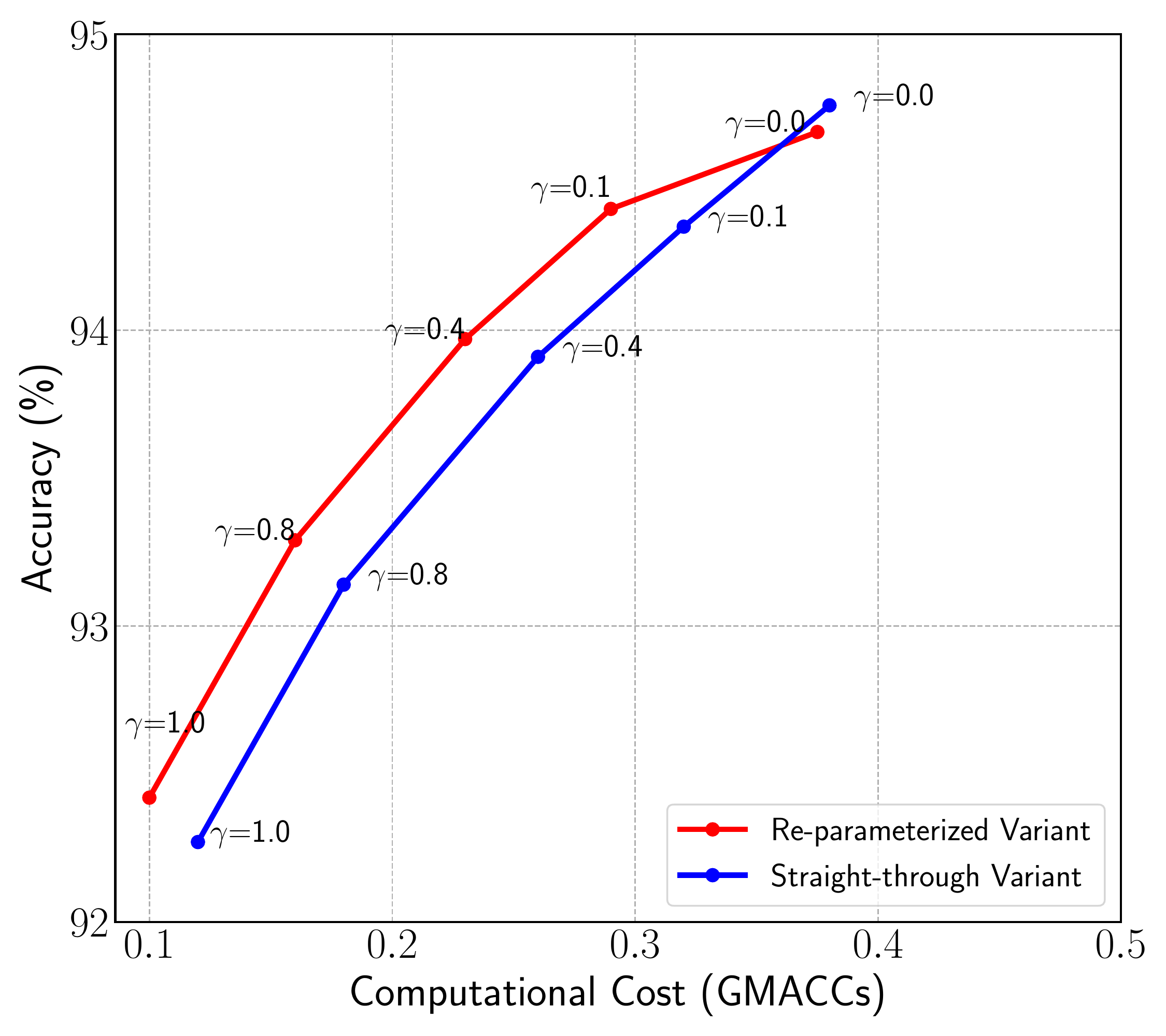}
		\label{sfg.gumbel_a}
	}
	\hspace{-1.6em}
	\subfigure[CIFAR-100]{
		\centering
		\includegraphics[width=.487\linewidth]{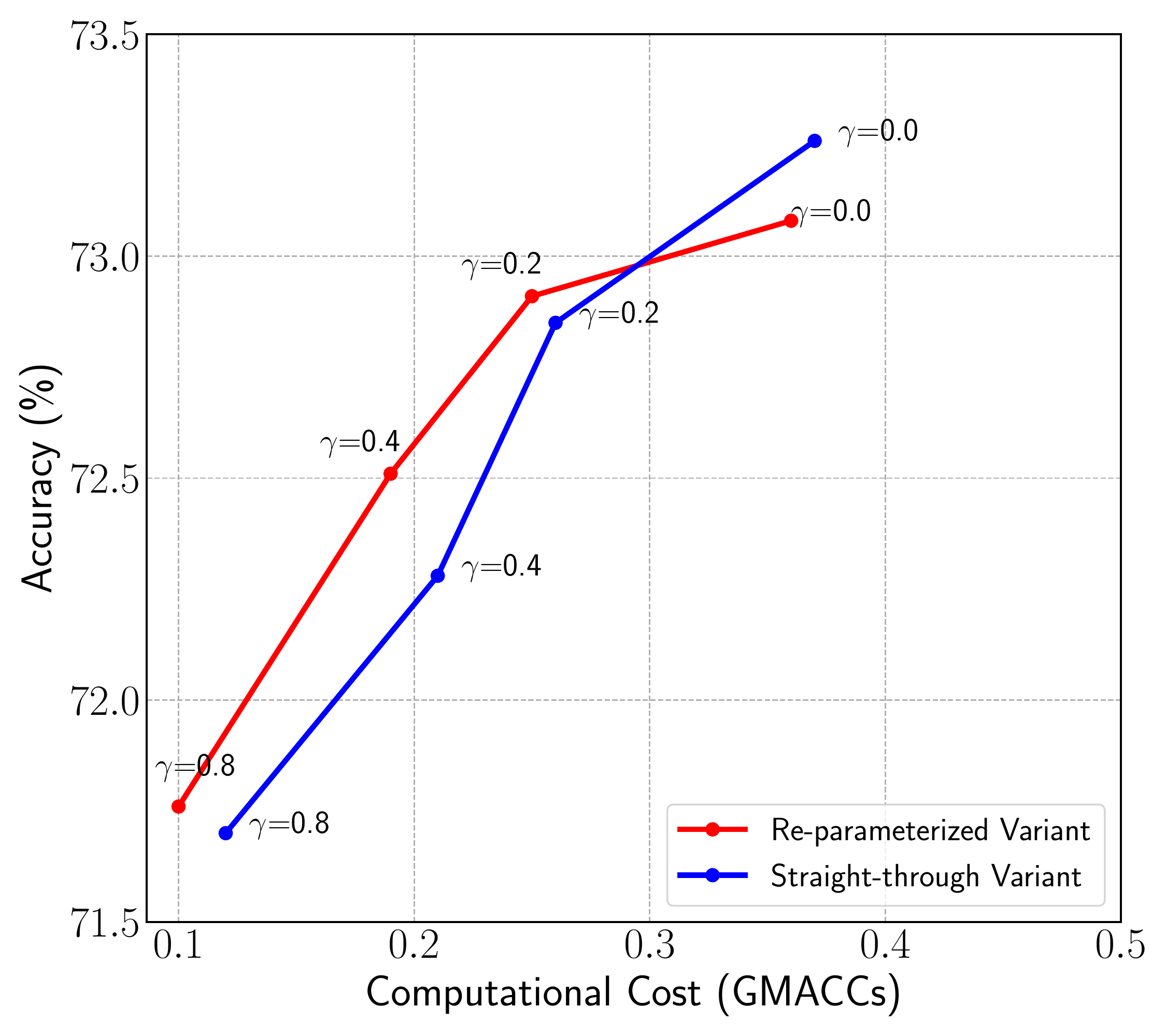}
		\label{sfg.gumbel_b}
	}
	\caption{The accuracy against computational cost of the re-parameterized and the straight-through Gumbel-Softmax variants on CIFAR-10/100. Results are based on ResNet-110. Best viewed in color.}
	\label{fig.Gumbel_Softmax}
\end{figure}
\subsection{Effectiveness of Gumbel-Softmax}\label{gumbel_softmax} 
To train the dynamic routing network end-to-end, relaxation methods are employed because the binary routing paths are not differentiable. As discussed in \cref{sec.relaxation}, we adopt Gumbel-Softmax for in our method. In this section, we compare the two variants of Gumbel-Softmax, \ie, the re-parameterized variant and the straight-through variant. As shown in \cref{fig.Gumbel_Softmax}, the re-parameterized variant performs better than the straight-through variant in most cases on CIFAR-10 and CIFAR-100. Weighing the pros and cons, we take the re-parameterized variant Gumbel-Softmax in our method.

\begin{figure}[h]
	\centering
	\subfigure[CIFAR-10]{
		\centering
		\includegraphics[width=.487\linewidth]{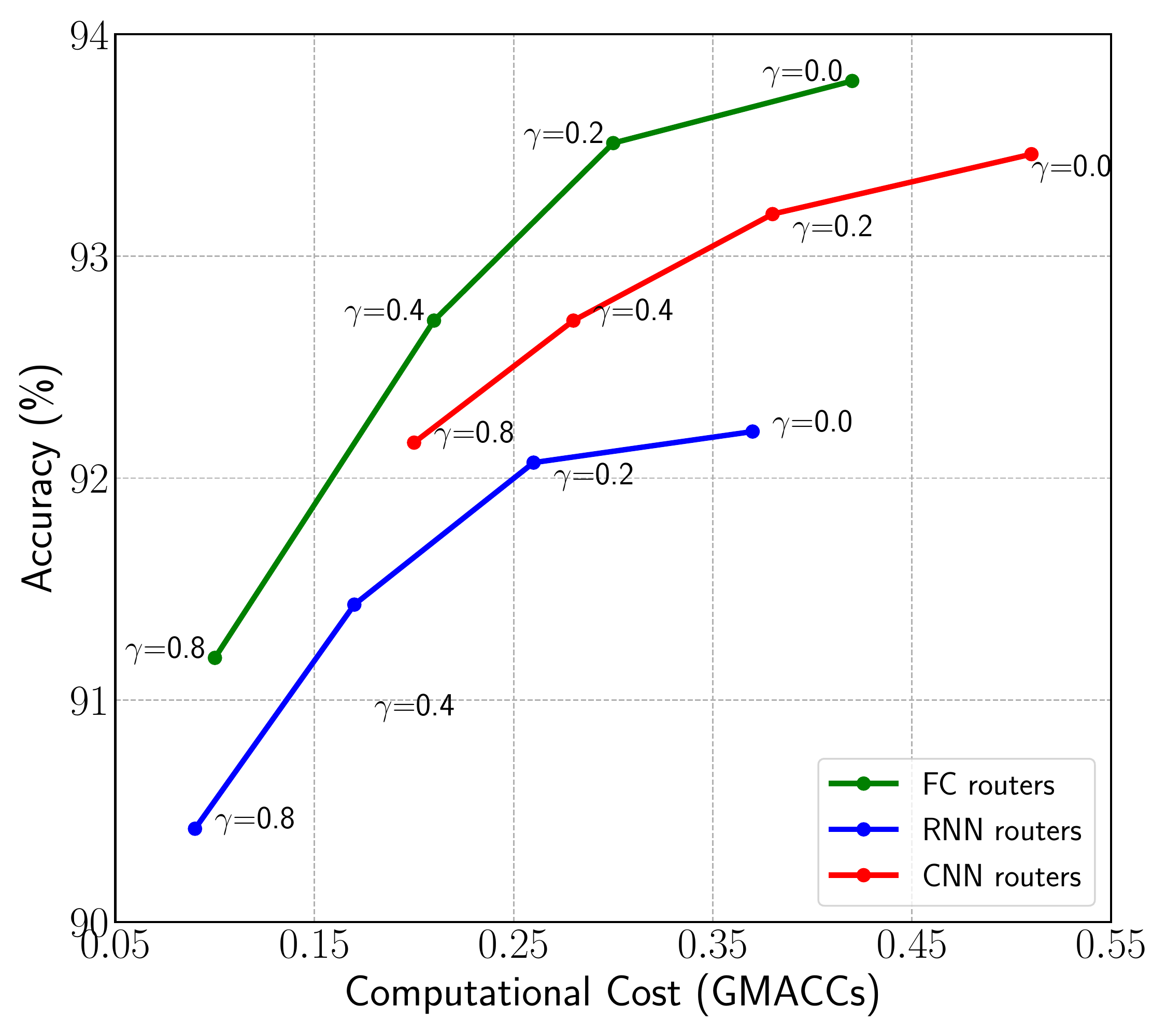}
		\label{sfg.router_a}
	}
	\hspace{-1.6em}
	\subfigure[CIFAR-100]{
		\centering
		\includegraphics[width=.487\linewidth]{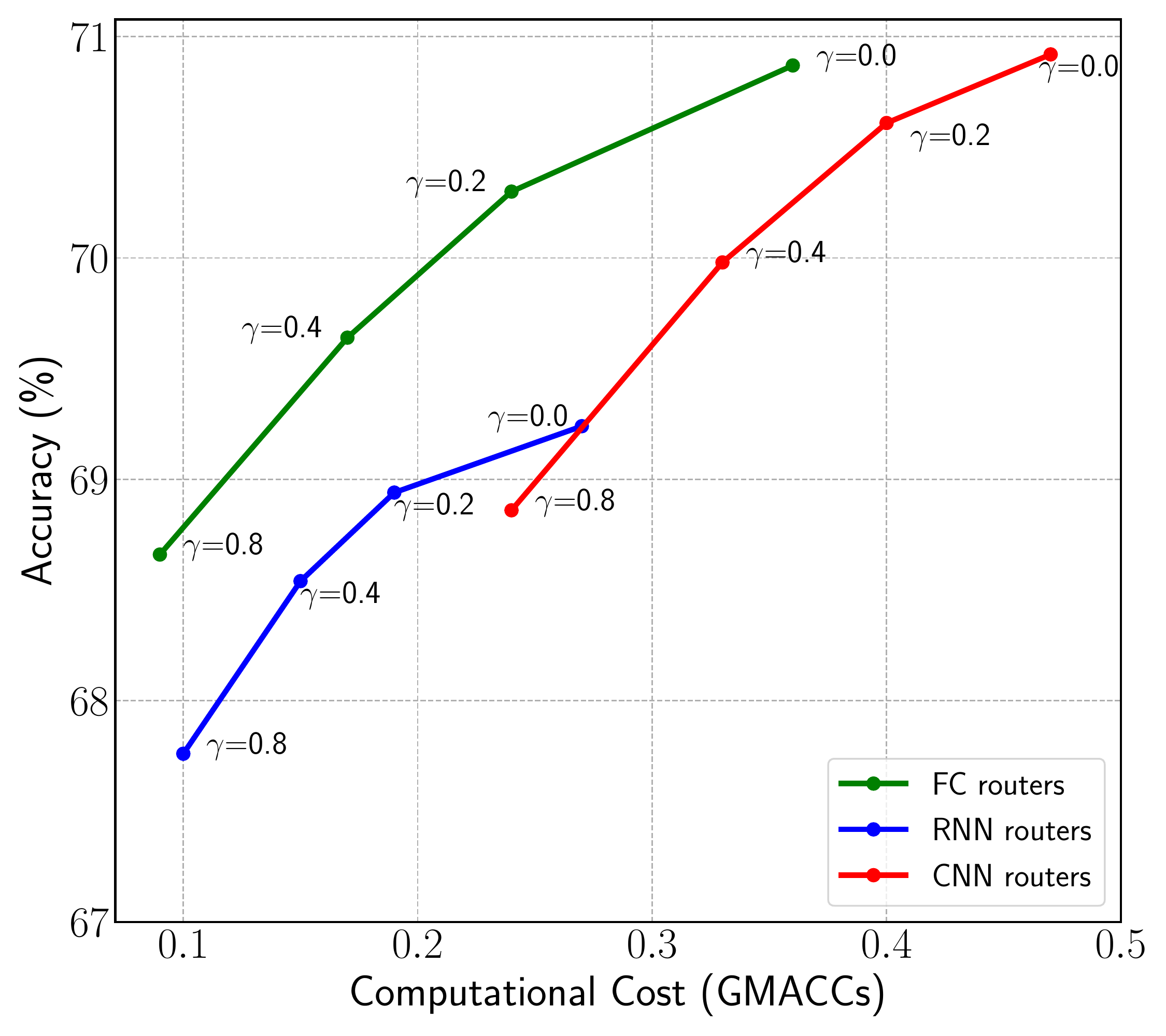}
		\label{sfg.router_b}
	}
	\caption{The accuracy against computational cost of the CNN router, the RNN router, and the FC router on CIFAR-10/100. Results are based on ResNet-110. Best viewed in color.}
	\label{fig.routers_compare}
\end{figure}
\subsection{Advantages of Our Router}\label{model_compare}
Routers are key components in a dynamic routing network, which make execution decision for blocks. How to design a lightweight yet effective router has always been the focus in dynamic routing. 
In this section, we discuss different types of routers: the CNN router, the RNN router, and the FC router. As shown in~\cref{fig.routers_compare}, we show the accuracy against computational cost, when a network equipped with different routers.

Specifically, a CNN router is composed of a $3\times 3$ convolutional layer followed by a global average pooling layer and a linear layer to output $1\times2$ vector. A RNN router is composed of a global average pooling, a shared linear layer, and a shared LSTM layer with a hidden unit size of 10. For the FC router, it uses two linear layers, after a global average pooling. Please refer to~\cite{veit2018convolutional, wang2018skipnet} for more details.
As a result, the FC router achieves the highest accuracy under multiple computational settings, comparing with the RNN router and the CNN router.


\end{document}